\documentclass[lettersize,journal]{IEEEtran}

\usepackage{stfloats}

\usepackage{graphicx} 
\usepackage{xcolor}
\usepackage{amsmath}
\usepackage{amssymb}

\usepackage{multirow}
\usepackage{graphicx}
\usepackage{tcolorbox}

\usepackage[caption=false,font=footnotesize]{subfig}
\usepackage{hyperref} 

\usepackage{algorithmicx}
\usepackage[ruled, vlined, linesnumbered, resetcount]{algorithm2e}
\SetKwInput{KwInput}{Input}                
\SetKwInput{KwOutput}{Output} 
\let\oldnl\nl
\newcommand{\nonl}{\renewcommand{\nl}{\let\nl\oldnl}}
\SetKwRepeat{Do}{do}{while}%
\def\HiLi{\leavevmode\rlap{\hbox to \hsize{\color{gray!35}\leaders\hrule height .8\baselineskip depth .5ex\hfill}}}
\def\HiLii{\leavevmode\rlap{\hbox to \hsize{\color{blue!10}\leaders\hrule height .8\baselineskip depth .5ex\hfill}}}

\usepackage{todonotes}
\newcommand{\Rev}[1]{\textcolor{black}{{#1}}}
\newcommand{\Revv}[1]{\textcolor{black}{{#1}}}

\newcommand{\TR}[2]{#2}
\newcommand{\TRR}[2]{#2}

\begin{document}

\title{A Highly Efficient Diversity-based Input Selection for DNN Improvement Using VLMs}

\author{Amin Abbasishahkoo,
        Mahboubeh Dadkhah,
        Lionel Briand,~\IEEEmembership{Fellow,~IEEE}
\thanks{Amin Abbasishahkoo is with the School of EECS, University of Ottawa,
Ottawa, ON K1N 6N5, Canada (e-mail: aabba038@uottawa.ca).}
\thanks{Mahboubeh Dadkhah is with the School of EECS, University of Ottawa,
Ottawa, ON K1N 6N5, Canada (e-mail: mdadkhah@uottawa.ca).}
\thanks{Lionel Briand is with the School of EECS, University of Ottawa, Ottawa, ON K1N 6N5, Canada, and also with the Research Ireland Lero centre for software, University of Limerick, Ireland (e-mail: lbriand@uottawa.ca).}

\thanks{Manuscript received March 1, 2025.}}

\maketitle
\begin{abstract}
Maintaining or improving the performance of Deep Neural Networks (DNNs) through fine-tuning requires labeling newly collected inputs, a process that is often costly and time-consuming. To alleviate this problem, input selection approaches have been developed in recent years to identify small, yet highly informative subsets for labeling. Diversity-based selection is one of the most effective approaches for this purpose. However, they are often computationally intensive and lack scalability for large input sets, limiting their practical applicability. To address this challenge, we introduce Concept-Based Diversity (CBD), a \TR{C3.3}{novel and highly efficient diversity} metric for image inputs that leverages Vision-Language Models (VLM). Our results show that CBD strongly correlates with Geometric Diversity (GD), an established diversity metric, while requiring only a fraction of the computation time. This makes CBD an efficient and scalable alternative for guiding repetitive, extensive input selection scenarios. Building on this finding, we propose a hybrid input selection approach that combines CBD with Margin, a simple uncertainty metric. We conduct a comprehensive evaluation across a diverse set of DNN models, input sets, selection budgets, and \Rev{six} most effective state-of-the-art selection baselines. The results show that CBD-based selection consistently outperforms all baselines in guiding input selection to improve the DNN model. Particularly, the CBD-based selection achieves an average accuracy improvement percentage of 38\% across various
budget sizes for ImageNet, which corresponds to a \Rev{5.7} percentage points increase relative to the second-best performing baselines.
Furthermore, the CBD-based selection approach remains highly efficient, requiring selection times close to those of simple uncertainty-based methods such as Margin, even on larger input sets like ImageNet. These results confirm not only the effectiveness and computational advantage of the CBD-based approach, particularly compared to hybrid baselines, but also its scalability in repetitive and extensive input selection scenarios.


\end{abstract}

\IEEEpeerreviewmaketitle


%




\section{Introduction}
\label{sec:Introduction}

Deep Neural Networks (DNNs) have been used in a wide range of application domains, including computer vision, health, and particularly software engineering~\cite{chai2021deep, yang2022survey, shahkoo2023autonomous}. 
Similar to traditional software, DNNs
require continuous updates to maintain and improve their performance, particularly as new inputs become available. 
However, manually labeling all newly collected inputs, whether for testing or fine-tuning, is often impractical because it requires substantial time and effort. Input selection addresses this challenge by identifying and labeling only a small, yet highly informative subset of the candidate inputs. Subsequently, fine-tuning the DNN on this carefully selected subset may enable the model to achieve performance comparable to training on the full set of new inputs while significantly reducing labeling costs.
In recent years, several input selection methods have been proposed, mostly by the software engineering community~\cite{hu2024test}, aiming to improve DNN model performance while reducing fine-tuning costs.
Diversity-based selection approaches are widely used and highly effective for this purpose; however, they tend to be computationally expensive, especially when dealing with large unlabeled input sets in practice. To overcome this limitation, we introduce a highly efficient Concept-Based Diversity (CBD) metric for measuring the diversity of image input sets, leveraging Vision-Language Models (VLM). Building on this metric, we further present a hybrid input selection approach that leverages both diversity and uncertainty to guide fine-tuning of vision DNNs (DNNs that process images as inputs), providing not only effective but also efficient model improvement.

VLMs are deep learning models that
learn joint visual–textual representations through training on massive collections of image-text pairs.
\TR{C3.9}{The proposed CBD metric leverages VLMs, such as CLIP (Contrastive Language-Image Pretraining)~\cite{radford2021learning} and SigLIP (Sigmoid Loss for Language-Image Pretraining)~\cite{zhai2023siglip}, which comprise an image encoder and a text encoder that jointly map their respective inputs to a shared multimodal embedding space.} This shared space enables us to calculate the similarity between an image embedding and a set of text embeddings, and extract the most representative concepts for each image. Subsequently, the CBD score for an image input set is calculated based on the diversity of the concepts extracted from images in that set. \Rev{To alleviate the overhead of running the VLM for each image input, we adopt a recent technique (described in the subsequent section) to map an image representation from the DNN model's representation to the shared embedding space of the selected VLM, enabling highly efficient CBD calculation while remaining applicable to different VLMs, including both CLIP and SigLIP.}


Diversity metrics have been successfully used for input selection both in traditional software and in DNN models~\cite{aghababaeyan2024deepgd, aghababaeyan2023black}. 
A recent comprehensive study on diversity metrics for image input sets, 
showed that Geometric Diversity (GD) is the most effective metric in capturing the actual diversity of an image input set~\cite{aghababaeyan2023black}. 
Despite its demonstrated effectiveness, measuring diversity efficiently remains a major concern, particularly for metrics that require pairwise distance calculations over large input sets, which can become computationally expensive, and for input selection methods that require measuring diversity across a large number of candidate subsets.

In contrast, the proposed CBD metric measures the diversity of an input set based on the natural-language concepts that represent the images in the set, effectively and efficiently. The efficiency of the CBD score calculation, enabled by leveraging VLMs, is particularly important in the context of input selection, as in such scenarios, a large number of candidate input subsets are often examined until the final input subset is selected. Compared to well-established input diversity metrics such as GD, CBD provides a comparable diversity measurement effectiveness, with 
significantly less computation time. Similar to GD, the CBD score is independent of the DNN model's characteristics; thus, it can reliably measure input diversity even when the model has low accuracy. 
\TR{C3.3, C2.1}{Furthermore, beyond advancing diversity estimation by reformulating it into highly efficient computation in the concept space, the proposed CBD metric is inherently domain-agnostic. Its core idea, calculating the diversity of an input set based on semantic concepts extracted from its inputs, can be generalized to other modalities by leveraging corresponding multimodal foundation models.}

In addition to diversity metrics, uncertainty metrics, including Margin~\cite{hu2024test}, can also be leveraged for selection. 
While these metrics are highly efficient and effective at detecting DNN failures, they are less effective in guiding model improvement. This is because they often prioritize inputs near the decision boundary, where many inputs share similar characteristics, leading to redundancy among selected inputs. In contrast, diversity metrics help identify input sets with more diverse, non-redundant inputs.
However, diversity-based approaches might select inputs that are likely to be correctly predicted by the model and thus not contribute to model improvement. 

To address the limitations of using either uncertainty or diversity alone, hybrid selection methods have been proposed that combine both of these metrics~\cite{sun2023robust, aghababaeyan2024deepgd}. Although Hybrid selection approaches are generally more effective at guiding model improvement, they are often computationally expensive because they require calculating the diversity through pairwise distances in the high-dimensional feature space of many candidate input subsets before final selection. In this paper, we propose a hybrid input selection approach based on our CBD metric that not only consistently outperforms state-of-the-art (SOTA) baselines in guiding DNN model improvement but also remains highly efficient. 


We empirically evaluated our proposed CBD metric in measuring the actual diversity of input sets. \TR{C3.9}{We calculated the CBD scores using two leading VLMs, CLIP and SigLIP, in our evaluations.} Our results on \Rev{three} subject models and input sets, across various subset sizes (1\%, 3\%, 5\%, 7\%, and 10\% of each original test set), demonstrate that our proposed CBD score, \Rev{calculated using both CLIP and SigLIP,} has a strong correlation with the GD score, an established diversity metric, indicating its high potential to guide DNN model improvement. \Rev{In addition to its effectiveness in measuring diversity, this finding also demonstrates that the CBD metric is VLM-agnostic.}
Furthermore, CBD is highly efficient, with calculation time approximately 2.5 to \Rev{35} times faster than GD while delivering comparable diversity measurement performance.


Moreover, we performed a comprehensive study to evaluate our CBD-based input selection and compared it with \Rev{six} SOTA baselines. Given the study's extensive scope and high computational cost, we limited our comparison to the most effective and computationally feasible selection approaches.
Conducting our experiments took \Revv{approximately 61 days (excluding interruptions).}
This substantial computational cost stems from several factors: the large number of baselines, subjects, and selection budgets considered; the time-intensive process of fine-tuning DNN models using selected input subsets; and the need to repeat the fine-tuning process five times for each combination of baseline, subject, and selection budget to account for the inherent stochasticity of DNN fine-tuning.

The results demonstrate that our CBD-based selection approach consistently outperforms baselines in improving the DNN model across all subjects and selection budgets through fine-tuning.
We also compared the time required to select subsets with the same selection budgets for the CBD-based approach and all the investigated baselines.
Notably, the CBD-based selection approach remains highly efficient, requiring computation time comparable to that of simple uncertainty-based methods such as Margin. Moreover, compared with other hybrid and diversity-based selection baselines, the CBD-based approach reduces selection time by more than three orders of magnitude, making it practical and highly scalable, even for larger input sets such as ImageNet~\cite{ILSVRC15}. 

\TR{C3.9}{We also investigated the effect of parameter configuration, i.e., the initial uncertainty-based subset selection size, on the performance of the proposed CBD-based selection approach. The results show that while the approach maintains competitive performance across a broad range of parameter values, initializing the selection with a relatively small uncertainty-based subset (between 0.05 and 0.2 of the selection budget) provides the best balance between uncertainty and diversity, enabling the CBD metric to more effectively refine the remaining selected inputs and maximize DNN improvement.}
The key contributions of this paper are as follows:
\begin{itemize}
    \item A \TR{C3.3}{novel, highly efficient} metric for measuring the diversity of image input sets using VLMs, named CBD. Our empirical evaluation shows that CBD is not only effective at measuring the actual diversity of an input set but is also highly efficient. Consequently, CBD is a practical and scalable metric for repetitive and extensive input selection from larger input sets such as ImageNet. \item \TR{C3.9}{We show that CBD is VLM-agnostic, i.e.,  it can be calculated using various VLMs such as CLIP or SIGLIP to measure diversity of image input sets. Furthermore, CBD has the potential to be applied to inputs in other modalities by leveraging multi-modal foundation models that provide shared embeddings for the target modality and text.}
    \item A \TR{C3.3}{new hybrid} input selection approach capable of effectively guiding DNN model improvement while featuring competitive efficiency. The proposed approach combines CBD and uncertainty scores to select a small yet diverse and informative subset for effectively fine-tuning the DNN model.
    \item An extensive evaluation that took \Revv{approximately 61 days} to complete, comparing the proposed CBD-based selection with \Rev{six} SOTA baselines across \Rev{three} subject DNN models and input sets, including the large ImageNet. The results indicate that the CBD-based approach consistently outperforms all baselines, achieving significantly greater improvements in model accuracy when fine-tuning with subsets selected under the same selection budget.  
    \item \TR{C1.7, C2.3}{A comprehensive parameter sensitivity analysis to identify the optimal configuration of the proposed hybrid approach for balancing uncertainty and diversity, maximizing the effectiveness of input selection for retraining the DNN model.}
    \item \TRR{C3.5}{An ablation study that confirms the contribution of both the uncertainty-based and diversity-based selection components of the proposed CBD-based approach.}
\end{itemize}

The remainder of this paper is organized as follows: Section II provides background, including a brief overview of the input selection baselines used in our experiments and a summary of VLMs and concept extraction from images. Section III introduces the proposed CBD metric and presents the CBD-based approach for input selection. Section IV describes the experimental design used to evaluate the CBD metric and the CBD-based selection approach. Section V reports and discusses the results for each research question. Section VI reviews related work, and Section VII concludes the paper.





\section{Background}
\label{sec:Background}
This section provides an overview of the main categories of SOTA input selection approaches, used as baselines in our evaluations.
Additionally, since we introduce a new concept-based method for calculating the diversity of an input set that leverages VLMs, we also review VLMs and discuss how they can be utilized to automatically extract concepts from image inputs.

\subsection{Input Selection for DNN improvement}
\label{sec:BackgroundInputSelection}
This section reviews related work on input selection approaches for enhancing DNN models, which we consider baselines in our evaluations. To make our experiments feasible in terms of computational resources, we intentionally exclude certain existing input selection approaches based on their reported effectiveness or computational cost. Specifically, we exclude approaches that have been comprehensively studied and evaluated in prior work and that are outperformed by some baselines included in our experiments. 

For instance, Adaptive Test Selection (ATS)~\cite{gao2022adaptive} introduced by Gao \textit{et al.} has consistently been outperformed by more recent baselines~\cite{aghababaeyan2024deepgd, hu2024test}. ATS iteratively selects inputs based on the dissimilarity between each remaining input and the currently selected subset, making it one of the most computationally intensive selection approaches~\cite{hu2024test}. Similarly, Surprise Adequacy (SA) metrics, proposed by Kim \textit{et al.}~\cite{kim2019guiding}, and Neuron coverage (NC) metrics~\cite{pei2017deepxplore, Ma2018DeepGaugeMT} are among the white-box methods excluded for their limited effectiveness. SA metrics measure how surprising an input is relative to the DNN model's training input set, while NC metrics consider the neuron activation ranges. Both ATS and coverage-based metrics have also been consistently outperformed by more recent selection approaches in terms of improving DNN performance~\cite{gao2022adaptive, hu2024test, aghababaeyan2024deepgd, sun2023robust}.




\subsubsection{\textbf{Uncertainty metrics}}
The primary objective of selection approaches based on uncertainty metrics is to identify and select inputs for which the DNN model is less confident in its prediction~\cite{feng2020deepgini}. Consequently, they tend to select inputs that are close to the decision boundary. Since these methods focus on uncertainty rather than diversity, the selected inputs might not cover a wide or representative region of the DNN’s input space. However, some of these methods have proven effective not only for detecting DNN faults but also for guiding model improvement~\cite{hu2024test, Weiss2022SimpleTechniques, feng2020deepgini, li2024distance}. Therefore, in this section, we review uncertainty-based methods that have been shown to effectively guide model improvement. 

Uncertainty metrics measure a DNN model's confidence in predicting an input, either by analyzing the model's predicted probability distribution~\cite{hu2024test} or by leveraging information from the input's nearest neighbors in the input set. DeepGini~\cite{feng2020deepgini}, Margin~\cite{hu2024test}, and Vanilla-Softmax~\cite{Weiss2022SimpleTechniques} are widely known metrics that rely solely on the model's predicted probability for the given input. However, DeepGini and Vanilla-Softmax have been outperformed by other baselines~\cite{aghababaeyan2024deepgd, li2024distance, sun2023robust} in terms of improving the accuracy of DNN models. Therefore, we exclude them from our experiments. 
The Margin score quantifies a model’s confidence in distinguishing its top two predicted classes for an input $x$, defined as the difference between the highest ($p_m$) and second-highest ($p_n$) predicted probabilities: $\text{Margin (x)} = p_m(x) - p_n(x)$.




Recent research has investigated using a given input's neighbors to estimate the model's uncertainty in predicting it~\cite{li2024distance, bao2023defense}. One of the most effective methods in this category is DATIS (Distance-Aware Test Input Selection), proposed by Li \textit{et al.}~\cite{li2024distance}, which outperforms prior input selection techniques in both fault detection and model performance improvement~\cite{abbasishahkoo2025metasel, li2024distance}. Unlike uncertainty metrics that rely solely on predicted probabilities, DATIS estimates uncertainty based on the ground-truth labels of the input’s nearest neighbors within the DNN model's labeled training set. First, it calculates the support of the training inputs for the DNN's prediction for a given input $x$. This support is computed by summing the exponentials of the distances between the given input and its closest training neighbors. Specifically, the support of the training inputs for predicting an input ${x}$ as part of class $c \in Y$ based on its $k$-nearest neighbors in the training set is calculated as follows:

\begin{equation}
p^*_c (x) = \frac{\sum_{t \in Train_k(x)} \exp\left(-\|z(x) - z(t)\|^2_2/ \tau\right) \mathbb{I}(y(t) = c)}
{\sum_{t \in Train_k(x)} \exp\left(-\|z(x) - z(t)\|^2_2/ \tau\right)}
\end{equation}

\noindent where $Train_k(x)$ is the list of $x$'s $k$-nearest neighbors in the training set, $\tau$ is a scaling parameter controlling the influence of distances in the latent space, and $z(x)$ and $z(t)$ are the latent feature representations of the given input $x$ and its nearest neighbor $t$, respectively. 
$\mathbb{I}(.)$ represents an indicator function that equals 1 if the predicted label of the training input $y(t)$ matches class $c$, and 0 otherwise.  After calculating $p^*_c (x)$ for each class $c \in Y$, an estimation $p^*(x) = \{p^*_1, p^*_2, \dots, p^*_C\}$ for the input $x$ is obtained. Then the DATIS uncertainty score for $x$ is calculated as $\text{DATIS}(x) = p^*_n / p^*_m$.


\noindent where $m$ is the class predicted by the DNN model for input $x$, which may or may not correspond to the class with the highest support $p^*_c (x)$. 
$n  =\arg\max_{c \in \mathcal{Y}, c \neq m} p^*(x)$ denotes the most supported prediction distinct from the DNN predicted class $m$. Therefore, a higher value of $\text{DATIS}(x)$ indicates that the DNN’s prediction for $x$ is weakly supported by the training inputs.



\subsubsection{\textbf{Diversity metrics}}
\label{sec:DiversityMetrics}
Unlike uncertainty metrics, diversity metrics measure how well the inputs in an input set collectively cover different regions of the input or feature space. 
Aghababaeyan \textit{et al.}~\cite{aghababaeyan2023black} conducted a comprehensive study on black-box, model-agnostic diversity metrics for image inputs, including Normalized Compression Distance, Standard Deviation, and GD. These metrics measure diversity without requiring access to the model's internals. Because they are independent of the DNN model, they can reliably measure the diversity of an input set even when the model's accuracy is low. Their results show that GD best captures the true diversity of an input set and, thus, selects inputs to test and improve DNN models~\cite{aghababaeyan2023black}. They also show that GD correlates statistically with fault detection in DNNs and outperforms other evaluated diversity metrics~\cite{aghababaeyan2023black}. Building on this finding, they proposed DeepGD~\cite{aghababaeyan2024deepgd}, a hybrid search-based approach that integrates GD with DeepGini for input selection.
Another metric, introduced by Gao \textit{et al.}~\cite{gao2022adaptive}, measures the dissimilarity between an input and a candidate set using local fault patterns defined by clustering the model's outputs. Building on this metric, they propose an adaptive input selection method, ATS, to select a diverse set of inputs that can effectively detect faults and improve the DNN model's accuracy.

Despite their effectiveness, a major limitation of most diversity-based selection methods lies in their substantial computational cost. Specifically, methods that calculate pairwise distances between inputs or measure the dissimilarity between each input and a selected subset are among the most expensive and inefficient input selection methods~\cite{gao2022adaptive, aghababaeyan2023black, hu2024test}. Furthermore, this cost grows substantially with the size of the input set, thereby limiting the scalability of these methods in scenarios requiring extensive testing and making them even more impractical for larger input sets. For instance, Hu \textit{et al.}~\cite{hu2025assessing} showed that deploying ATS incurs substantial computational overhead, particularly on larger input sets, where its execution time grows substantially compared to other already demanding approaches.




\subsubsection{\textbf{Hybrid methods}}
Hybrid input selection approaches aim to leverage the complementary strengths of multiple criteria to enhance effectiveness. 
In particular, several approaches combine uncertainty and diversity metrics~\cite{sun2023robust, aghababaeyan2024deepgd}. Approaches that rely solely on uncertainty metrics are generally efficient but tend to select inputs near the decision boundary, often leading to redundant inputs that contribute little to improving the DNN model.
In contrast, approaches that focus solely on diversity aim to select inputs that differ in the feature space, thereby reducing redundancy. However, they might select inputs the DNN model is likely to predict correctly and, therefore, less effective at improving the model.
By combining these two perspectives, hybrid approaches balance both aspects to select more effective and representative input subsets.


For instance, Aghababaeyan \textit{et al.} developed DeepGD~\cite{aghababaeyan2024deepgd}, an input selection approach that combines both GD and DeepGini metrics. To achieve this, their multi-objective search-based approach deploys the Non-dominated Sorting Genetic Algorithm (NSGA-II).
Their comprehensive evaluations demonstrate that DeepGD achieves superior performance not only in detecting a diverse range of faults but also in effectively guiding input selection to improve DNN performance.
Even with small selection budgets, DeepGD-selected input subsets consistently and significantly improve the model's accuracy compared with uncertainty-based, coverage-based, and other diversity-based approaches such as ATS~\cite{gao2022adaptive}. However, as discussed in the following sections, DeepGD's high computational overhead makes it expensive in practice.

Wang \textit{et al.}~\cite{wang2025sets} address DeepGD's computational limitation by introducing SETS (Simple yet Effective Test Selection), a hybrid two-phase selection approach that minimizes costly diversity calculations. SETS proposes a lightweight, greedy algorithm that combines uncertainty and diversity for input selection. In the reduction phase, it filters out low-uncertainty inputs, retaining only those more likely to detect faults. The number of retained inputs is controlled by a user-defined parameter $\alpha$, which determines the proportion of inputs preserved for selection. In the original implementation, $\alpha$ is set to retain three times the final selection budget (i.e., $\alpha=3*b$), focusing only on the most uncertain inputs.
Then, in the selection phase, the reduced candidate set is partitioned into equidistant sub-groups based on sorted uncertainty, ensuring even coverage across uncertainty levels. Rather than evaluating all candidates in each selection step, it selects only one input from each sub-group using a fitness function that combines uncertainty and normalized diversity gain. This reduces the total number of diversity evaluations from quadratic (in standard greedy selection) to linear in the number of candidate inputs, significantly reducing diversity computation cost.

While SETS~\cite{wang2025sets} achieves substantial efficiency gains by aggressively limiting the number of diversity calculations, this restriction noticeably affects its effectiveness, as our results in the following sections show.
By filtering out a large portion of low-uncertainty inputs in the early reduction phase, SETS significantly lowers computational cost, particularly for large input sets. However, this early pruning can reduce the quality of the final selected subset, especially for small $\alpha$, because it may discard potentially valuable inputs that could detect diverse faults or improve the model.
Moreover, in the selection phase, SETS further constrains diversity by selecting only one input from each subgroup, which may exclude potential inputs and further limit retraining effectiveness in certain settings.

Robust Test Selection (RTS), proposed by Sun \textit{et al.}~\cite{sun2023robust}, is another hybrid input selection method that combines uncertainty and diversity to prioritize inputs. RTS divides unlabeled candidate inputs into three distinct sets: noise, suspicious, and successful (correctly classified) based on the model's output probability distributions. Then it selects sequentially from these sets in a fixed priority order: first from suspicious, then from successful, and finally from noise inputs. To prioritize inputs in each set, RTS uses a probability-tier-matrix-based metric. As a result, RTS not only focuses on the diversity of the selected inputs, but it also avoids prioritizing inputs that are likely to be correctly predicted by the model. Their comprehensive evaluations demonstrate RTS's superior effectiveness in both fault detection and model improvement compared to uncertainty-based, coverage-based, and surprise-based selection approaches~\cite{sun2023robust}. 

\TR{C2.5}{In a recent study, Chen \textit{et al.}~\cite{Chen2026DiscrepantFeatures} introduced the Graph-Based Test Selection (GTS) that implements a three-step workflow. In the first step, redundant features in the DNN's latent space are identified and filtered out to improve the performance of feature-based similarity calculation in subsequent steps. 
Specifically, GTS constructs a KNN graph over the latent feature representations of unlabeled inputs and employs a Graph Convolution Network (GCN) to identify and remove redundant features shared by both high-confidence and low-confidence inputs. In the second step, it reconstructs the KNN graph solely from the refined features to predict a label for a given unlabeled input based on its nearest neighbors, and prioritizes those whose predicted labels differ from the DNN's original predictions. Finally, GTS removes noisy and highly redundant inputs, creating a compact and diverse subset with enhanced fault detection capability.
}

\subsection{Concept extraction from image inputs using VLMs}

\subsubsection{Vision Language Models}
Recent advancements in artificial intelligence have led to the emergence of VLMs, a new category of models that learn joint representations from both visual and textual inputs. Unlike traditional vision models that rely on manually labeled input sets with fixed labels, VLMs are trained on massive collections of image-text pairs. This large-scale, multimodal training enables them to generalize across diverse tasks and domains and, more importantly, to represent and describe visual concepts in natural language.

CLIP~\cite{radford2021learning}, proposed by OpenAI, is a well-known VLM that consists of two jointly trained encoders: an image encoder (such as a ResNet or Vision Transformer) and a text encoder (a Transformer model). Each encoder maps its input to a shared multimodal embedding space. Training is conducted using a contrastive learning objective on a dataset of 400 million image-text pairs extracted from the Internet. During the training phase, the model learns to maximize the similarity between matching image and text representations while minimizing the similarity between mismatched ones. 
\TR{C3.9}{SigLIP (Sigmoid Loss for Language-Image Pre-Training), introduced by Zhai \textit{et al.}~\cite{zhai2023siglip}, replaces CLIP’s global softmax normalization with a pairwise sigmoid loss. SigLIP is also trained on an Internet-based input set comprising up to 10 billion image-text pairs.
}

\begin{figure}
    \includegraphics[width = \columnwidth]{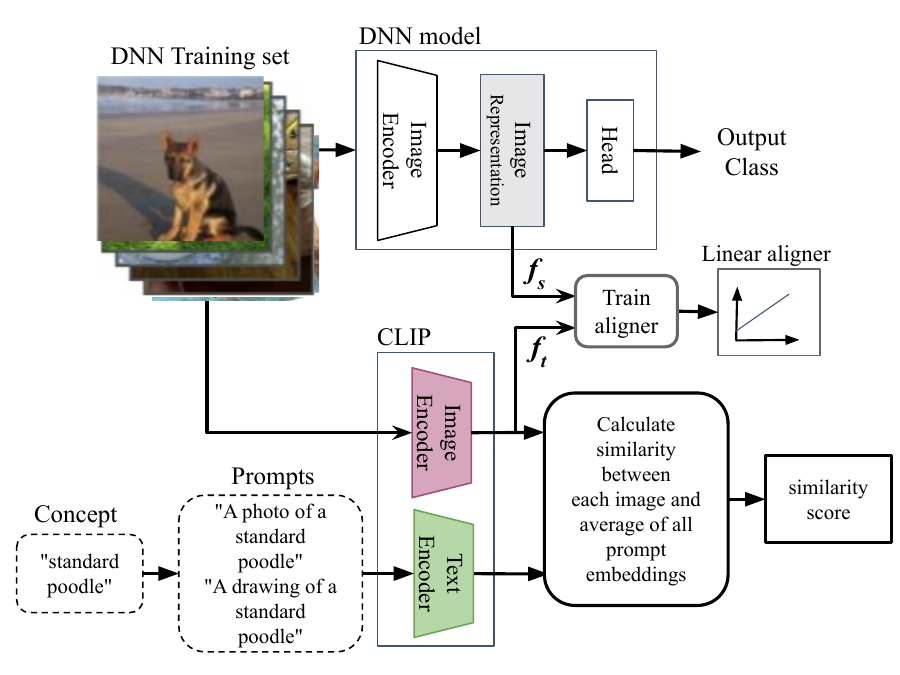}
    \centering
    \caption{Calculating similarity score between an image and a natural language concept using CLIP (lower part), and training a linear aligner between a DNN's internal representation and CLIP's shared embedding space (upper part).}
    \label{fig:background}
\end{figure}

In inference, \Rev{VLMs such as CLIP and SigLIP can} perform zero-shot classification by comparing an image's embedding with those of a set of text prompts containing all ground-truth labels of an input set (e.g., "a photo of a car", "a photo of a cat"), and selecting the one with the highest similarity score. Similarly, it can be used for extracting related concepts from image inputs. To achieve this, an image's embedding can be compared with those of a set of text prompts containing a diverse set of related concepts in the underlying input domain. Subsequently, the most similar concepts can represent the image. 
For instance, consider a general input set containing images of different animals and a general knowledge base, such as ConceptNet~\cite{speer2017conceptnet}. As shown in the lower part of Figure~\ref{fig:background}, for each concept, such as "standard poodle", we create a set of prompts that include it. For example, consider using the following prompt templates:
\begin{itemize}
    \item "A photo of a" + concept
    \item "A drawing of a" + concept
\end{itemize}

For example, for the concept "standard poodle", two prompts are constructed: "A photo of a standard poodle" and "A drawing of a standard poodle". Then, each prompt is processed by the \Rev{VLM's text encoder (e.g., CLIP or SigLIP)} to obtain its corresponding embedding, and the average of these two embeddings is calculated to represent a single embedding for each concept~\cite{radford2021learning, moayeri2023text}.
Similarly, we create two prompts for each concept in the knowledge base and average their embeddings to form a diverse set of concept embeddings.
Given an image, the VLM's image encoder processes it to obtain an embedding, which we then compare against all concept embeddings. Consequently, the concepts whose corresponding embeddings achieve the highest similarity scores are selected as the ones representing that image.


\subsubsection{Knowledge bases for concept extraction}
\label{sec:BackgroundKnowledgeBases}
To extract concepts from an input set of images, we require a knowledge base containing a diverse set of related concepts in the underlying application domain. A variety of general taxonomies and ontologies are available, including WordNet~\cite{miller1995wordnet}, ConceptNet~\cite{speer2017conceptnet}, as well as more specialized domain-specific ontologies. Beyond general-purpose knowledge bases, specific visual concept sets support vision-based tasks, such as the Visual Genome scene graph~\cite{krishna2017visual} and the Open Images dataset~\cite{kuznetsova2020open}. 
Visual Genome is a large-scale dataset with annotated scene graphs that represent visual concepts as objects, attributes, and relationships in images. This dataset defines more than 100,000 object categories and 40,000 attribute categories, along with spatial and functional relationships (e.g., on top of, holding, wearing). The resulting structure is a graph of visual entities, where nodes represent objects or attributes, and edges capture pairwise relationships. Unlike purely linguistic resources such as WordNet or ConceptNet, the Visual Genome ontology is grounded in visual semantics, making it highly relevant for vision-based tasks and for bridging the gap between image content and conceptual representation.



In our experiments, we consider two main criteria for selecting a knowledge base: 1) to contain visual concepts, i.e., concepts that represent entities or objects that can be observed in an image, and 2) to contain concepts corresponding to the application domain of our subject input sets. Since we ran experiments on general image input sets such as ImageNet, as outlined in subsequent sections, we leverage the Visual Genome scene graph~\cite{krishna2017visual}, which represents general visual concepts. It is important to note that while our proposed CBD metric is not limited to general domain applications, applying it to a domain-specific input set requires selecting an appropriate knowledge base that captures diverse domain concepts.


\subsubsection{DNN to VLM image representation alignment}
\label{sec:alignment}

To extract related concepts from an input image using \Rev{a VLM (such as CLIP or SigLIP)}, each input image must be processed by the \Rev{VLM}'s image encoder to obtain its corresponding embedding in the shared space. As a result, in an input selection scenario with an extensive set of candidate inputs, the \Rev{VLM} must run once per candidate, creating significant computational overhead. In particular, for hybrid input selection approaches that leverage both the DNN model's output for uncertainty calculation and the \Rev{VLM}'s similarity scores for diversity calculation, eliminating the need for repeated \Rev{VLM} executions can substantially improve efficiency.
Moayeri \textit{et al.}~\cite{moayeri2023text} show that a model's image representation can be mapped to another model's representation using a simple linear alignment.  
The task of mapping between a source vision encoder $f_s$ and a target vision encoder $f_t$ (i.e., $h : f_s(\mathcal{X}) \rightarrow f_t(\mathcal{X})$) can be learned by 
minimizing the mean squared error between the target encoder’s embedding of an image and the source encoder’s embedding after applying a trained linear transformation. In other words, this mapping is achieved by solving the following optimization problem:
\begin{equation}
W, b = \arg \min_{W,b} \; \frac{1}{|D_{\text{train}}|} 
\sum_{x \in D_{\text{train}}} 
\| W^T f_s(x) + b - f_t(x) \|_2^2.
\end{equation}

\noindent where $D_{\text{train}}$ represents the training set and a vision encoder is a model $f$ that maps images $x \in \mathcal{X}$ to vectors $f(x) \in \mathbb{R}^d$. The mapping $h$ is also restricted to the class of affine transformations (i.e., $h_{W,b}(z):= W^T z + b$).
They investigated widely used vision models, such as the ResNet series and CLIP~\cite{radford2021learning}, and reported that such mappings can be learned surprisingly well with just a single linear layer~\cite{moayeri2023text}. This finding suggests that, in input-selection approaches where the DNN model's input-image representations are already available, a linear aligner can map those representations to \Rev{the VLM}'s shared embedding space, as illustrated in the upper part of Figure~\ref{fig:background}. Consequently, we can obtain \Rev{VLM}'s corresponding embeddings without running it on every new input, substantially reducing computational cost and improving efficiency. In our input selection approach described in Section~\ref{sec:Approach}, we build and use a linear aligner for each DNN model to efficiently extract concepts for each image, without running \Rev{the VLM} directly.

\section{Approach}
\label{sec:Approach}
In this section, we first introduce our concept-based diversity (CBD) metric for image input sets, calculated by analyzing the concepts that represent the images in the set. To investigate its efficiency and effectiveness in guiding input selection to improve the DNN model, we present a hybrid selection approach leveraging CBD.


\subsection{Concept-based diversity metric}
\label{sec:ConceptBasedDiversityApproach}
The overall process of calculating the proposed CBD metric includes two main steps. The first step, presented in Algorithm~\ref{Alg:CRSConstructionAlignerTraining}, initializes the model and needs to be performed only once for each DNN and its corresponding training set. The second step, presented in Algorithm~\ref{Alg:CBDScoreCalculation}, calculates the diversity of a given unlabeled input set and can be applied multiple times to many input sets.


%

\begin{algorithm}[t!]
\DontPrintSemicolon
  \KwInput{\;
    \Indp $N$: the DNN model \; 
    $TrainSet$: the labeled training input set of $N$ \;
    $ImgEncoder_{VLM}$: \Rev{VLM}'s image encoder \;
    $TxtEncoder_{VLM}$: \Rev{VLM}'s text encoder \; 
    $KNB$: Knowledge base \; 
    }
    
    \KwOutput{ \;
    \Indp $RCS$: Representative Concept Set \;
    $Aligner_N$: Trained linear aligner between $N$ and \Rev{VLM} spaces\;
    } 

    \nonl
    \HiLi 1) Train a linear aligner between $N$ and \Rev{VLM} spaces \;
    \ForEach{image $img_i$ in $TrainSet$}{
        $fs_i \xleftarrow{} GetDNNRepresentation(img_i);$ \;
        $ft_i \xleftarrow{} GetVLMEmbedding(img_i);$ \;
    }
    $Aligner_N \xleftarrow{} TrainLinearAligner(fs, ft);$ \;

    \nonl
    \HiLi 2) Construct RCS \;
    $ConceptList \xleftarrow{} LoadConcepts(KNB);$ \;
    \nonl
    \HiLii Extract each concept's \Rev{VLM} embedding \;
    \ForEach{concept $cnp_j$ in $ConceptList$}{
        $PmtList \xleftarrow{} CreatePrompts(cnp_i);$ \;
        $EmbList \xleftarrow{} GetVLMEmbedding(PmtList);$ \;
        $fe_j \xleftarrow{} Average(EmbList);$ \;
    }
    \nonl
    \HiLii Extract each image's top $m$ concepts \;
    \ForEach{image embedding $ft_i$ in $ft$}{
        \ForEach{concept embedding $fe_j$ in $fe$}{
            $SmtList \xleftarrow{} Similarity(ft_i, fe_j);$ \;
        }
        $topConcepts \xleftarrow{} TopConcepts(SmtList, m);$ \;
        $A \xleftarrow{} AddtoArchive(topConcepts);$ \;
    }
    \nonl
    \HiLii Select unique concepts \;
    $RCS \xleftarrow{} RemoveDuplicates(A);$ \;
    \nonl
    \HiLii Save RCS concepts corresponding \Rev{VLM} embeddings \;
    $RCSemb \xleftarrow{} SaveEmbeddings($fe, RCS$);  $ \;

    \KwRet{$RCSemb, Aligner_N$}\; 
    
\caption{Constructing RCS and training a linear aligner between a  DNN and \Rev{VLM} embedding spaces}
\label{Alg:CRSConstructionAlignerTraining}
\end{algorithm}



\subsubsection{\textbf{Initialization step}}

This initial step has two objectives: train a linear aligner to map the DNN model's image representation to its embedding in \Rev{VLM}'s embedding space, and construct a set of human-interpretable concepts that span the concept space on which the DNN model has been trained. 
To achieve the former, as described in Algorithm~\ref{Alg:CRSConstructionAlignerTraining}, each training image $img_i$ is fed into both the DNN model and \Rev{the target VLM} to obtain the DNN's internal representation and \Rev{VLM} embeddings, denoted as $fs_i$ and $ft_i$, respectively (lines 1-3). Using these paired embeddings for every training image, an accurate linear aligner can be built from the DNN representation space to the \Rev{shared VLM} space (line 4). 
This aligner plays a crucial role in the efficiency of the CBD calculation, enabling fast, accurate concept extraction for unseen, unlabeled inputs in subsequent steps. By leveraging the DNN’s representation of an input and mapping it to \Rev{VLM}’s embedding space using the aligner, concept extraction can be performed without directly invoking \Rev{VLM} for every new input, significantly reducing computational cost. In Section~\ref{sec:AlignerExperiments}, we report the accuracy of the aligners we build and investigate the efficiency improvements of using the aligner compared to directly extracting concepts with \Rev{VLM}.
The results show that an accurate aligner can be built for each DNN model, enabling efficient and accurate concept extraction.

In the initialization step, a set of highly related concepts, called the Representative Concept Set (RCS), is also created from the images in the training set, collectively representing the semantic concept space on which the DNN model has been trained. While this process has traditionally required extensive manual effort to annotate images, the introduction of VLMs enables automatic concept extraction. To achieve this, we rely on existing knowledge bases that include concepts describing the underlying domain knowledge specific to each input set, as described in Section~\ref{sec:BackgroundKnowledgeBases}.


First, all concepts are retrieved from the selected knowledge base (line 5).
Next, for each concept in the knowledge base, we create a set of natural language prompts and encode them using \Rev{VLM}'s text encoder (lines 7 and 8). The most widely used prompt template for image inputs follows the pattern “A photo of [concept]”. However, specialized prompt templates can be adapted or extended depending on the input set and application domain to achieve more precise concept representations. The complete set of prompt templates used for each input set in our experiments is available in our replication package~\cite{replicationpackage}. We adopt the process of using multiple prompt templates for each concept as proposed in the original paper~\cite{radford2021learning} and widely adopted in subsequent works~\cite{moayeri2023text}, to mitigate \Rev{VLM}’s sensitivity to prompt phrasing and boost its performance. As a result of this process, we obtain multiple embeddings for each concept based on its associated prompts. Then, we average these embedding vectors to obtain a single embedding for each concept~\cite{radford2021learning, moayeri2023text} (line 9).  



Now that we have obtained the embeddings for all concepts $fe$, and the corresponding embeddings of the training images in \Rev{VLM}'s shared embedding space $ft$, previously extracted during linear aligner training, we calculate the similarity score between each image and all concept embeddings (lines 10-12). We then select the top $m$ most similar concepts for each image and add them to an archive list of all top concepts across the training set (lines  13 and 14). Finally, we remove duplicate concepts from the list, and the resulting set of unique, highly related concepts forms the RCS of the input set (line 15).
The embeddings of all RCS concepts in \Rev{VLM}’s shared space are stored for the subsequent step of our approach, where they are utilized to calculate the diversity score of an input set (line 16). The goal of creating RCS is not only to obtain a set of concepts that represent the specific concept space of each input set, but also to improve CBD calculation efficiency by preventing similarity calculations between each new unlabeled image and unrelated concepts in the knowledge base. As a result, to extract concepts representing each unlabeled image in the input selection step, we calculate similarity only to the concepts in RCS, rather than to all concepts in the knowledge base.

\begin{algorithm}[t!]
\DontPrintSemicolon
  \KwInput{\;
    \Indp $N$: the DNN model \; 
    $RCS$: Representative Concept Set \; 
    $Aligner_N$: Trained linear aligner between $N$ and \Rev{VLM} spaces\; 
    $TS$: Unlabeled input set \;
    }
    
    \KwOutput{ \;
    \Indp $CBD$: CBD score of input set $TS$ \;
    } 

    \nonl
    \HiLii Load RCS concepts corresponding embeddings \Rev{in shared VLM space} \;
    $RCSemb \xleftarrow{} LoadRCSEmbeddings(RCS);$ \;
    
    \nonl
    \HiLii Map image embeddings with the trained linear aligner \;
    \ForEach{image $I_i$ in $TS$}{
        $fs_i \xleftarrow{} GetDNNRepresentation(I_i);$ \;
        $ft_i \xleftarrow{} Map(Aligner_N, fs_i);$ \;
    }

    \nonl
    \HiLii Calculate similarity of an image to RCS concepts \;
    \ForEach{image $ft_i$ in $ft$}{
        \ForEach{concept $fe_j$ in $RCSemb$}{
            $SmtList \xleftarrow{} Similarity(ft_i, fe_j);$ \;
        }
        $topConcepts_i \xleftarrow{} TopConcepts(SmtList, m);$ \;
    }

    \nonl
    \HiLii Calculate CBD \;
    \ForEach{concept $con_c$ in $topConcepts$}{        
        $fq(con_c) \xleftarrow{}$ frequency of $con_c$ across all concepts in $topConcepts$ in \; 
    }
    $totalFreq \xleftarrow{} \sum_{j=1}^{k} fq(con_j)$ \;
    \ForEach{concept $con_c$ in $topConcepts$}{        
        $p_c \xleftarrow{} \frac{fq(con_c)}{totalFreq}$ \;
    }
    
   $CBD(TS) \xleftarrow{} - \sum_{j=1}^{k} p_j \log_2(p_j)$ \;

    \KwRet{$CBD (TS)$}\; 
    
\caption{CBD-score calculation for an unlabeled input subset TS}
\label{Alg:CBDScoreCalculation}
\end{algorithm}

\begin{figure}
    \includegraphics[width = \columnwidth]{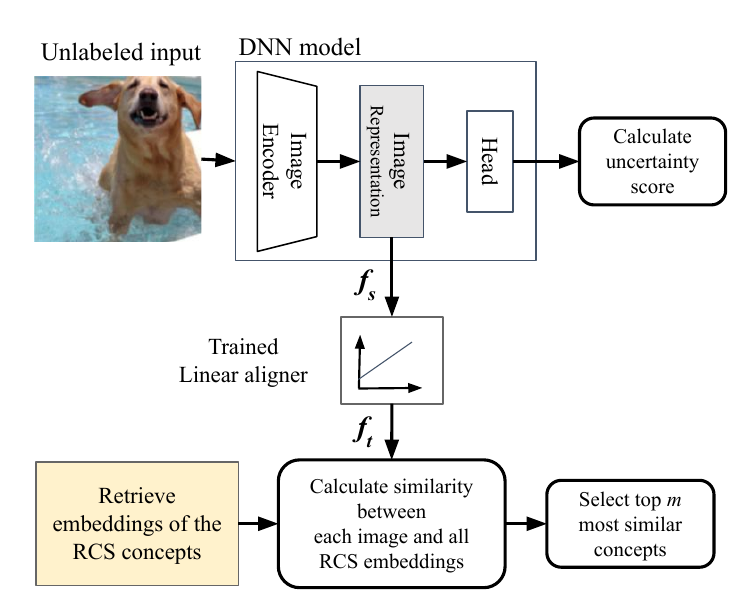}
    \centering
    \caption{The efficient process of finding highly similar concepts for each image using the RCS and the trained linear aligner.}
    \label{fig:ConceptExtraction}
\end{figure}

\subsubsection{\textbf{CBD score calculation step}}

In this step, we efficiently calculate the CBD score of a given unlabeled image input set using the RCS and the aligner constructed in the previous step, without running CLIP, as described in Algorithm~\ref{Alg:CBDScoreCalculation}. First, the embeddings of the concepts in the RCS obtained and stored during the initialization step are retrieved (line 1). Then, the DNN processes each image in the unlabeled input set, extracts its internal representation, and maps it to CLIP's shared embedding space using the linear aligner constructed in the previous step (lines 2-4). Subsequently, the similarity scores are computed between each mapped image embedding and all RCS concept embeddings (lines 5-7). Finally, for each image, we select the top $m$ most similar concepts to represent its semantic content (line 8), which serves as the basis for calculating the concept-based diversity score.

Figure~\ref{fig:ConceptExtraction} illustrates the efficient process we used to identify the most similar concepts from the RCS for each input image. As shown, we first process each image with the DNN to compute an uncertainty score from the model's output probabilities, while concurrently extracting its representation in the DNN's internal space. The trained linear aligner then maps this representation into CLIP’s shared embedding space, eliminating the need to execute CLIP’s image encoder. Moreover, we extract and store the embeddings of all RCS concepts in the CLIP embedding space during initialization (Algorithm~\ref{Alg:CRSConstructionAlignerTraining}) for reuse in this phase, avoiding repeated executions of CLIP’s text encoder. As a result, the computational overhead of concept extraction is limited to a forward pass through the linear aligner and similarity calculations between image and concept embeddings.
We also discuss the efficiency of CBD score calculation and elaborate on this process in Section~\ref{sec:AlignerExperiments}.

To calculate the diversity of an input set based on concepts extracted from its input images, we use the Shannon Entropy (lines 9-14). Let $TS = \{I_1, I_2, \dots, I_b\}$ denote the selected input set with size $b$, where each input image $I_i$ is represented by a set of $m$ most similar concepts $topConcepts(I_i)= \{Con_{1}, Con_{2}, \dots, Con_{m}\}$. While optimizing $m$ enhances the ability of the CBD score to capture the true diversity of an input subset and thereby improves selection effectiveness, we set $m=10$ in our experiments using a lightweight manual procedure. Specifically, for each input set, we sampled a representative subset of images and manually analyzed the concepts with the highest similarity scores. Based on this inspection, we identified a cutoff point beyond which the remaining concepts were weakly related to the input images and largely constituted noise.
A comprehensive optimization of $m$ would require extensive additional experiments, which is computationally infeasible given the already high cost of our current experiments. Therefore, we leave a more thorough investigation of this parameter as future work.
In our experiments, as discussed in the following sections, we set $m=10$. 
Let $fq(Con_{c})$ denote the frequency of concept $Con_{c}$ across the entire input set $TS$. The probability of observing a concept $Con_{c}$ is then defined as:

\begin{equation}
p_c = \frac{fq(Con_{c})}{\sum_{j=1}^{k} fq(Con_{j})}
\end{equation}

\noindent where $k$ is the total number of distinct concepts observed in $TS$. Then, we calculate the CBD score of $TS$ as the Shannon entropy of the input set's concept distribution as follows:

\begin{equation}
CBD(TS) = - \sum_{j=1}^{k} p_j \log_2(p_j)
\end{equation}

\noindent This measures how uniformly the input set spans its related concepts. A higher entropy value reflects a more balanced distribution of concepts, indicating greater diversity. 

\subsection{CBD-based input selection}
\label{sec:CBD-basedSelectionApproach}

In this section, we combine our proposed CBD metric with simple uncertainty metrics to introduce a hybrid approach to input selection. Our goal is to develop a selection approach that is not only competitive with SOTA baselines in guiding DNN model improvement but also remains highly efficient.
Evaluating this CBD-based input selection approach enables us to investigate both the effectiveness and scalability of the CBD metric in the context of input selection, where the diversity of a large number of candidate subsets must be computed repeatedly.


\begin{algorithm}[t!]
\DontPrintSemicolon
  \KwInput{\;
    \Indp $N$: the DNN model \; 
    $CandidateSet$: the unlabeled candidate set for input selection \;
    $b$: Selection budget \;
    }
    
    \KwOutput{ \;
    \Indp $Selected$: Selected input subset \;
    } 

    \nonl
    \HiLi 1) Rank all inputs based on their uncertainty score  \;
    \ForEach{input $x_i$ in $CandidateSet$}{
        $U_i \xleftarrow{} CalculateUncertainty(x_i);$ \;
    }
    $Candidates \xleftarrow{} SortDescend(CandidateSet, U);$ \;

    \nonl
    \HiLi 2) Select a diverse subset based on CBD score \;
    \nonl
    \HiLii Select an initial set of inputs \;
    $Selected \xleftarrow{} null; $ \;
    $b_{init} \xleftarrow{} b/10;$ \;    
    $j \xleftarrow{} 0; $ \;
    \While{$j \neq b_{init}$}{
        $Concat(Selected, Candidates_j);$ \;
        $j \xleftarrow{} j + 1;$ \;
    }

    \nonl
    \HiLii Continue selection with non-redundant inputs \;
    \While{$j \neq b$}{
        $Temp \xleftarrow{} Concat(Selected, Candidates_j);$ \;
        \If{$CBDScore(Temp) > CBDScore(Selected)$}{
            $Selected \xleftarrow{} Temp;$ \;
        }
        $j \xleftarrow{} j + 1;$ \;
    }
    
    \KwRet{$Selected$}\;     
\caption{CBD-based input selection approach}
\label{Alg:CBD-basedSelection}
\end{algorithm}

To combine the CBD score with input uncertainty, the proposed CBD-based input selection approach, as described in Algorithm~\ref{Alg:CBD-basedSelection}, has two main steps.
First, all inputs are ranked in descending order of their uncertainty scores (lines 1-3) to prioritize selecting inputs for which the DNN model exhibits low confidence. 
These are the most informative inputs, and fine-tuning the model with them can effectively enhance its performance. However, many of these inputs are redundant and thus unable to contribute to model improvement. To address this, in the second step, we use the CBD score to detect and eliminate redundant inputs, ensuring the final selected subset remains diverse.

To achieve this, we start by selecting a small initial subset of the most uncertain inputs, corresponding to a fraction of the selection budget (lines 4–8). In our experiments, for a selection budget of size $b$, we initialize the subset with $0.1b$ inputs. We then traverse the remaining inputs in descending order of uncertainty and evaluate them one by one based on their ability to increase diversity of the currently selected subset (lines 9-13). Specifically, for each candidate input, we temporarily add it to the current subset and compute the resulting CBD score. If this temporary subset achieves a higher CBD score than the current one, the input is accepted, and the selected subset is updated. Otherwise, the input is discarded, and the next candidate in the ordered list is investigated. This process continues until a final subset of size $b$ is selected.

By combining these two efficient and complementary criteria, the proposed CBD-based selection approach aims to select a subset that is both highly informative and diverse, enabling the fine-tuned model to achieve greater performance improvements within a constrained selection budget. 

\Rev{\subsection{Generalizability of the CBD metric and the proposed CBD-based selection approach}}

\TR{C2.1}{While our current implementation and evaluations focus on image input sets and using VLMs to extract representative textual concepts, the underlying core idea of the CBD metric, calculating the diversity of an input set based on its inputs' semantic concepts mapped from the model's latent space to human-interpretable language concept space using VLMs, is entirely domain-agnostic. Consequently, the CBD metric can be seamlessly adapted to estimate the diversity of input sets in other modalities by leveraging multi-modal foundation models that provide shared embeddings for the target modality and text. Subsequently, the proposed CBD-based selection approach can also be adapted to other domains. However, it is important to note that the proposed hybrid selection approach is specifically constrained to classification tasks. This is because the approach combines the CBD metric with a probability-based uncertainty metric, i.e., Margin, which requires class probability distributions to compute prediction uncertainty.
}


\section{Experiment Design}
\label{sec:ExperimentDesign}

\subsection{Research Questions}
\label{sec:ResearchQuestions}
\TR{C3.9}{We performed an extensive evaluation to answer five key questions. To answer the first two questions, which evaluate the effectiveness and efficiency of the proposed CBD metric, we ran experiments using two VLMs, CLIP and SigLIP, to show that CBD is VLM-agnostic and can be calculated effectively with different VLMs. Since both VLMs produced highly similar results, with CLIP exhibiting a slight performance advantage, we only used CLIP for the remaining questions to evaluate the performance of the CBD-based test selection approach. }
The research questions are defined as follows:

\textbf{\textit{RQ1: Does the proposed CBD metric effectively measure the diversity of an input subset?
}}
To answer this question, following related studies on diversity metrics~\cite{aghababaeyan2023black}, we can run a controlled experiment by creating many input subsets with systematically increasing actual diversity. Then, for each subset, we compute the concept-based diversity and observe how it changes as the actual diversity increases~\cite{aghababaeyan2023black}. In classification problems, a practical way to increase a subset's actual diversity is to start by randomly selecting all inputs from a single output class, then incrementally increase the number of covered classes by replacing some inputs with new ones sampled from different classes~\cite{aghababaeyan2023black}.

However, a more comprehensive way to evaluate our proposed CBD metric is to check whether it correlates with an established input diversity metric that, based on extensive study, aligns with actual diversity and strongly correlates with fault-detection capability~\cite{aghababaeyan2023black}. As discussed in Section~\ref{sec:DiversityMetrics}, extensive empirical studies have identified the GD score as the most effective diversity metric for DNNs, reliably measuring actual input diversity and showing a positive, statistically significant correlation with faults in DNNs~\cite{aghababaeyan2023black}. 
Consequently, this analysis allows us to evaluate the CBD metric's effectiveness for input selection and to efficiently guide DNN model improvement.

\textbf{\textit{RQ2: Is the proposed CBD metric efficient?
}}

To answer this question, we compare the cost of calculating an input subset's diversity using the proposed metric with that of the GD score~\cite{aghababaeyan2023black}, an established diversity metric for DNNs. As discussed in Section~\ref{sec:DiversityMetrics}, diversity metrics are among the most computationally expensive and inefficient input selection methods. Notably, their computational cost increases rapidly with larger input sets, often exponentially, making them impractical. Consequently, to compare computational efficiency, we measure the execution time to compute each metric on input subsets corresponding to 1\%, 3\%, 5\%, 7\%, and 10\% of the original test set, investigating each metric's scalability and efficiency on larger input sets.



\textbf{\textit{RQ3: Does the proposed CBD-based input selection approach guide DNN model improvement more effectively than baselines?
}}

To answer this question, we evaluate how effectively our input selection approach guides DNN model improvement and compare it with the baseline selection methods described in Section~\ref{sec:BackgroundInputSelection}. To do so, we compare the accuracy improvements from fine-tuning the DNN model with the input subset selected by the proposed approach and each baseline. We perform our evaluations across different selection budgets of 3\%, 5\%, 7\%, 10\%, 15\%, and 20\% of the original test set for each subject.

\textbf{\textit{RQ4: Is the proposed CBD-based input selection approach efficient?
}}

In this research question, we evaluate the efficiency of the proposed CBD-based selection approach by comparing its input selection time with that of the baselines. In particular, since diversity-based input selection methods are often among the most computationally expensive, our goal is to determine whether leveraging the proposed CBD metric can achieve not only high effectiveness in guiding input selection but also substantial improvements in computational efficiency.

\Rev{\textbf{\textit{RQ5: What is the optimal balance between uncertainty and diversity in the proposed hybrid approach? }}       }



\TR{C1.7, C2.3}{In this research question, we investigate the effect of the size of the initial selected subset of the most uncertain inputs ($b_{init}$) on the effectiveness of our proposed CBD-based input selection approach. 
While our default configuration initializes the selected subset with 0.1 of the selection budget ($b_{init} \xleftarrow{} b/10;$), we systematically vary this parameter to alternative values of $0.05b$, $0.2b$, $0.3b$, $0.5b$, and $b$ for all investigated selection budgets ($b$) of 3\%, 5\%, 7\%, 10\%, 15\%, and 20\% of the original test set to find the optimized configuration for achieving high effectiveness in improving the DNN model. 
}




\begin{table}[t]
    \centering   
    \footnotesize 
    \caption{Information about the input sets and DNN models}
    \resizebox{\columnwidth}{!}{
    \begin{tabular}{|c|      c|  c|  c|   }
    \hline 
    input  &\multirow{2}{*}{Model}   &Training set    &Test set \\ 
    set   &                        &size    &size   \\ \hline 
          CIFAR-10  &ResNet18  &50,000   & 10,000  \\  \hline
          ImageNet       &ResNet101  &1,281,167    &50,000              \\  \hline
          \Rev{CIFAR-100}       &\Rev{DenseNet121}  &\Rev{50,000}    &\Rev{10,000}           \\  \hline

    \end{tabular}
    }
    \label{tab:Subjects}
\end{table}

\subsection{Subjects}
\label{sec:Subjects}
We conducted our experiments using a collection of well-known image input sets, including CIFAR-10~\cite{krizhevsky2009learning}, and ImageNet~\cite{ILSVRC15} that have been utilized in numerous empirical studies on DNN testing~\cite{shen2020multiple, aghababaeyan2024deepgd, abbasishahkoo2024teasma, hu2022empirical, berend2020cats}. 
ImageNet is a widely recognized large-scale input set designed
for visual object recognition research in the field of computer
vision. It includes over 14 million labeled images across 1,000
object categories~\cite{ILSVRC15}. We leverage its most commonly used subset, ImageNet-1k, from the ILSVRC2012 competition, with
a training set of 1,281,167 images and a validation set of 50,000
images for evaluation. CIFAR-10 is another widely used input set that includes images from 10 classes (e.g., cats, dogs, airplanes, cars). CIFAR-10 contains 32x32 cropped colored images, while images in the ImageNet input set are 224x224 pixels and feature higher resolution.
\TR{C1.5, C3.8}{CIFAR-100 is another widely used benchmark dataset comprising 100 classes and a total of 60,000 images. Like CIFAR-10, it consists of 32×32 color images and is split into 50,000 training images and 10,000 test images. Both CIFAR-10 and CIFAR-100 are standard benchmarks commonly used in computer vision research.}

\Rev{Using these input sets, we trained three DNN models with ResNet18~\cite{he2016deep}, ResNet101~\cite{he2016deep}, and DenseNet121~\cite{huang2017densely} as detailed in Table~\ref{tab:Subjects}. }
To avoid selection bias, we reused the training and test sets defined in the original sources~\cite {krizhevsky2009learning, ILSVRC15}. In our experiments, we use the training set to create an RCS and train a linear aligner for each input set and DNN model. We subsequently use the test sets for input selection, model fine-tuning, and evaluation, as described in the next section.

\subsection{Linear aligner training}
\label{sec:AlignerExperiments}
As described in Section~\ref{sec:ConceptBasedDiversityApproach}, our proposed approach uses a linear aligner to map the DNN model's representation of an input to its representation in \Rev{VLM}'s embedding space. Although this aligner is not part of this paper's contribution, it is a component of the proposed approach and affects the results. Thus, in this section, we investigate the accuracy and the computational cost of the aligners we have built for each subject DNN model in our experiments. \TR{C3.9}{Since the subsequent evaluation of the proposed CBD metric is conducted using both CLIP and SigLIP, we report the performance of the corresponding linear aligners trained for each VLM.}

Table~\ref {tab: Aligners} presents detailed information on the trained aligners.
Initially, it is important to note that, as Moayeri \textit{et al.}~\cite{moayeri2023text} have shown in their original paper for mapping the representation of an image in a source model to that of a target model using a simple linear alignment, the differences in the representation sizes of the two models significantly impact the ability to train an accurate aligner. Specifically, when the source model's representation size is smaller than the target model's embedding size, building an accurate aligner becomes challenging, often resulting in reduced accuracy. Consequently, both the subject DNN architectures and the selected layer for representation extraction affect the achievable accuracy of the trained aligner. 

\TR{C1.5, C3.8, C3.9}{The trained linear aligners achieved consistently high accuracy across both evaluated VLMs. Specifically, when aligning with CLIP, the aligners achieved $R^2$ values of 0.83 for CIFAR-10 and ResNet18, 0.72 for ImageNet and ResNet101, and 0.84 for CIFAR-100 and DenseNet121. Similarly, when aligning with SigLIP, the aligners yielded  $R^2$ values of 0.78, 0.72, and 0.80 across the respective input sets and models.} These values are comparable to the 
$R^2$ values reported by Moayeri et al. \cite{moayeri2023text} for ImageNet and ResNet-based models. 
Moreover, the results presented in the next section indicate that this level of accuracy is sufficient to produce reliable CBD scores, showing that the linear aligners perform well enough to enable effective CBD-based input selection and model improvement.

To extract the representations, we used the outputs of the penultimate layer of the DNN models, as this layer captures the most informative and semantically meaningful features \cite{bengio2013representation}. \TR{C1.2}{This choice is further supported by the work of Moayeri \textit{et al.}~\cite{moayeri2023text}, who explicitly evaluated different source layers for aligning vision DNN representations with the CLIP embedding space. Their empirical results showed that representations extracted from the penultimate layer achieve the most accurate linear alignment and the highest coefficient of determination.}



\begin{table}[t]
    \centering   
    \caption{Information on the trained linear aligners, including each DNN's representation size, embedding size of the \Rev{VLM (including both CLIP and SigLIP)}, coefficient of determination ($R^2$), and the time (in seconds) required to extract the related concepts for the entire test set using \Rev{VLM} directly and using the trained aligner with the DNN models}
    \resizebox{\columnwidth}{!}{
    

\begin{tabular}{|c|          c|  c|   c|   }
    \hline 
         &CIFAR-10     &ImageNet    &\Rev{CIFAR-100}   \\ \hline
          DNN's representation size      &512   &2048   &\Rev{1024}    \\  \hline
          CLIP's embedding size     &512 &512 &512     \\  \hline
          SigLIP's embedding size  &\Rev{768} &\Rev{768} &\Rev{768}  \\  \hline
          $R^2$ (based on CLIP)     &0.83   &0.72   &\Rev{0.84}     \\  \hline 
          $R^2$ (based on SigLIP)     &\Rev{0.78}   &\Rev{0.72}   &\Rev{0.80}     \\  \hline 
          DNN's execution time  &3.5 s   &105 s  &\Rev{10 s} \\  \hline
          CLIP's execution time  &4 s  &36 s   &\Rev{4 s}  \\  \hline
          SigLIP's execution time  &\Rev{ 5 s}  &\Rev{ 47 s}   &\Rev{ 16 s}  \\  \hline

    \end{tabular}
    }
    \label{tab:Aligners}
\end{table}

In addition to evaluating the aligner's accuracy, we investigated and compared the time required to extract related concepts from unlabeled input images across the entire test set when using the DNN model's representations and the trained aligner, versus directly using \Rev{VLM}. We focus on the total execution time for concept extraction across the entire test set, because this step (the second step in Section~\ref{sec:ConceptBasedDiversityApproach}) directly influences the overall efficiency of an input selection strategy that leverages the CBD score. Measuring the cost of this step lets us assess how concept-extraction time affects both CBD computation efficiency and the overall effectiveness of CBD-based input selection when using either DNN model representations mapped with the aligner or \Rev{VLM} directly.

As reported in Table~\ref{tab:Aligners}, for smaller networks like ResNet18, the required time to execute the DNN model to extract test input representations and then apply the trained aligner (with negligible execution time) to map those representations to \Rev{VLM}'s embedding space is less than that of running \Rev{the VLM (either CLIP or SigLIP)} itself for concept extraction. In contrast, for deeper networks like ResNet101, running the DNN on all inputs takes longer than running \Rev{VLM} directly. These results may suggest that for such DNNs, it is more efficient to extract concepts directly using \Rev{a VLM}. 
However, it is important to emphasize that, in most input selection strategies, including our CBD-based selection approach, diversity is not the sole guiding metric~\cite{hu2024test, aghababaeyan2024deepgd}.

Input selection approaches usually leverage diversity scores together with other metrics, such as the model's uncertainty in predicting an input~\cite{aghababaeyan2024deepgd}, which inherently requires evaluating all unlabeled inputs against the DNN model. In such scenarios, the DNN can process an input to obtain its representation and estimate the model's uncertainty simultaneously, whereas running \Rev{VLM} would add extra cost. 
Therefore, since our proposed input selection approach leverages both the CBD and uncertainty metrics, we train and use a linear aligner to avoid the additional overhead of \Rev{VLM} execution. Nevertheless, when only the CBD score is required, we can calculate it directly using concepts extracted from \Rev{VLM}. The high accuracy of our trained aligners ensures that the resulting CBD scores closely approximate those computed directly using \Rev{VLMs}.

\par\vspace{-7pt}
\subsection{Data Availability}
The replication package for our studies will be shared upon the paper's acceptance~\cite{replicationpackage}.

\par\vspace{-7pt}
\section{Results}
\label{sec:Results}

In this section, we present the results related to our research questions and discuss their practical implications. 
All the results are available in our replication package~\cite{replicationpackage}.

%

\subsection{\textbf{RQ1: Effectiveness of the CBD score}}
\label{sec:ResultsRQ1}
To address this research question, we require a large number of input subsets with varying diversity. To create these subsets, we followed a controlled process similar to that adopted in~\cite{aghababaeyan2023black}, as described in Section~\ref{sec:ResearchQuestions}, to systematically create input subsets with increasing actual diversity. It is important to note that for input sets with a large number of output classes, such as ImageNet, which contains 1000 classes, we need to start the controlled subset selection from more than one class, even for the smallest selection budget (e.g., 1\% of the test set corresponds to 500 inputs). This is because each class has a relatively small number of inputs (only 50 per class in ImageNet). 

Therefore, we start by randomly selecting a limited number of classes (e.g., 10) and randomly sampling a subset of size $b$ from them. While keeping the subset size fixed at $b$, we gradually increase the number of classes represented in the subset by replacing some inputs with new ones from a new class. As the number of covered classes grows, the actual diversity of the subset increases, even though its size remains constant. 
At the initial stage and after each increment, we measured the CBD and GD scores for each subset, as described in Section~\ref{sec:ResearchQuestions}. 
Subsequently, we conducted a correlation analysis between the computed CBD score and the GD score of these input subsets. Specifically, we measured the correlation between CBD and GD using the nonparametric Spearman's rank correlation coefficient across all subsets, since we cannot assume linearity. \TR{C3.9}{We performed our experiments using two different VLMs, i.e., CLIP and SigLIP, separately.} 
Furthermore, we repeated this analysis for input subsets of different sizes corresponding to 1\%, 3\%, 5\%, 7\%, and 10\% of the original test set for each subject.
The correlation results for all subjects and subset sizes, \Rev{using both CLIP and SigLIP}, are reported in Table~\ref{tab:correlationAnalysis}.
\Rev{While the results obtained using both VLMs demonstrate a strong correlation between the CBD and GD scores, the correlations achieved with CLIP are generally higher and more consistent than SigLIP. This is mainly due to the SigLIP linear aligner's lower accuracy, as reported in Section~\ref{sec:AlignerExperiments}. 
Specifically, with CLIP, the CBD score consistently correlates strongly with the GD score, with a minimum Spearman coefficient of 0.90 across all subjects and subset sizes.} 
This finding not only validates the effectiveness of the CBD metric in accurately measuring the actual diversity of an input subset, \Rev{but also demonstrates that the proposed CBD metric is VLM-agnostic and can be calculated using different VLMs.} 

\begin{table}[t]
    \centering   
    \caption{Spearman correlation coefficients computed between concept-based diversity and GD}
    \resizebox{\columnwidth}{!}{
    \begin{tabular}{|c| c|       c|  c|  c| c|  c|  }
    \hline 
    & \multirow{2}{*}{Input set}    &\multicolumn{5}{c|}{Input subset size} \\ \cline{3-7}
           & &$b$=1\%      &$b$=3\%    &$b$=5\%    &$b$=7\%    &$b$=10\%  \\ \cline{1-7}
        \multirow{3}{*}{CLIP}  &CIFAR-10    &0.90   &0.95  &0.95 &0.95  &0.91 \\  \cline{2-7}
          &ImageNet         &0.92    &0.92  &0.93      &0.94  &0.96      \\  \cline{2-7}
          &\Rev{CIFAR-100}         &\Rev{0.97}    &\Rev{0.97}  &\Rev{0.98}      &\Rev{0.98}  &\Rev{0.96}      \\  \hline

          \multirow{3}{*}{\Rev{SigLIP}}  &\Rev{CIFAR-10}    &\Rev{0.82}  &\Rev{0.82}  &\Rev{0.94} &\Rev{0.94}  &\Rev{0.83} \\  \cline{2-7}
          &\Rev{ImageNet}         &\Rev{0.90}    &\Rev{0.92}  &\Rev{0.93}     &\Rev{0.94}  &\Rev{0.94}      \\  \cline{2-7}
          &\Rev{CIFAR-100}         &\Rev{0.94}    &\Rev{0.82}  &\Rev{0.91}      &\Rev{0.86}  &\Rev{0.76}      \\  \hline
    \end{tabular}
    }
    \label{tab:correlationAnalysis}
\end{table}

\begin{tcolorbox}   
    \textbf{Answer to RQ1:} The CBD scores have a strong Spearman correlation with GD scores, an established diversity metric known to align closely with the actual diversity of an input subset. 
    This strong correlation indicates that CBD can effectively guide input selection to improve the DNN model.
    \Rev{Moreover, these strong correlations hold consistently across two leading VLMs (CLIP and SigLIP), demonstrating that the proposed CBD metric is inherently VLM-agnostic.}
\end{tcolorbox}

\subsection{\textbf{RQ2: Efficiency of the CBD score calculation}}
\label{sec:ResultsRQ2}
To answer this research question, we evaluate the computation time required to calculate CBD scores and compare it with the time required to calculate GD scores. We report the computation time for input subsets used in the correlation analysis experiment designed to answer the previous question. 
When measuring the calculation time for concept-based diversity, we consider only the time required to perform the second step of its calculation. 
\TR{C2.4, C3.4}{As explained in Section~\ref{sec:ConceptBasedDiversityApproach}, the initialization step, building a linear aligner and creating the RCS, is performed only once for each input set and DNN model. Once the aligner and RCS are built, they can be used multiple times to calculate the CBD scores for many unlabeled subsets. 
Therefore, the cost of the second step (Algorithm~\ref{Alg:CBDScoreCalculation}) reflects the practical computational cost of deploying the CBD metric during input selection, which we discuss in detail and compare with other input selection baselines in Section~\ref{sec:ResultsRQ4}.}

\begin{table}[t]
    \centering   
    \caption{Average diversity calculation time (milliseconds) for a single subset of different sizes using CBD and GD}
    \resizebox{\columnwidth}{!}{
    \begin{tabular}{|c|      c|  c|  c|  c|  c|  c|  }
    \hline 
    \multirow{2}{*}{Input set} &\multirow{2}{*}{Metric}   &\multicolumn{5}{c|}{Input subset size} \\ \cline{3-7}
        &   &$b$=1\%      &$b$=3\%    &$b$=5\%   &$b$=7\%    &$b$=10\% \\ \hline
          \multirow{2}{*}{CIFAR-10}  &CBD  &0.3	&0.7	&1.1	&1.53	&2.2     \\  \cline{2-7}
                                     &GD   &1.8	 &2.5	&3.1	&5.3	&14.6 \\  \hline
          
          \multirow{2}{*}{ImageNet}  &CBD  &1.4	&4.2  &6.9	&9.4	&12.5 \\  \cline{2-7}
                                     &GD   &3.7	 &27.4	&96.7	&203.3	&437.9 \\  \hline
        \multirow{2}{*}{\Rev{CIFAR-100}}  &\Rev{CBD}    &\Rev{0.3}   &\Rev{0.7}    &\Rev{1.3}   &\Rev{1.7}   &\Rev{2.3}   \\  \cline{2-7}
                                     &\Rev{GD}    &\Rev{2.3}    &\Rev{4.5}   &\Rev{5.7}   &\Rev{7.2}   &\Rev{15.9}    \\  \hline
    \end{tabular}
    }
    \label{tab:diversityCalculationTime}
\end{table}

\TR{C3.9}{It is important to note that since the CBD score is calculated using the trained linear aligner rather than directly running the VLM, the runtime of the CBD calculation is independent of the underlying VLM. Consequently, CBD calculation time remains identical regardless of the VLM used.}
Table~\ref{tab:diversityCalculationTime} reports the results.
The reported results confirm the computational efficiency of the proposed CBD metric. Across all input sets and subset sizes, calculating CBD scores consistently requires significantly less time than calculating GD scores. The reported computation times correspond to the average time required to calculate the diversity score for a single subset, whereas in most input selection strategies, such as search-based or adaptive random testing, diversity scores must be calculated repeatedly across many candidate subsets. Consequently, the execution time difference between CBD and GD becomes practically significant in the context of input selection.


As expected, the computation time for both metrics increases with the size of the input subset. However, this increase is substantially greater for the GD metric than for CBD. 
These results highlight the substantial computational advantage of CBD, especially for larger input sets such as ImageNet. In such scenarios, the subset selection budget is larger, and the selection process must compute diversity for many more candidate subsets, further increasing the computational cost of input selection. While both metrics are affected by both the input set size and the DNN model depth and complexity, CBD exhibits far superior scalability, maintaining its efficiency across different DNN architectures and input set sizes.

These findings, together with those from the first research question, demonstrate that CBD provides a level of diversity measurement effectiveness comparable to the GD metric at a fraction of its computational cost, making it a highly practical metric for large-scale input selection scenarios.




\begin{tcolorbox}   
    \textbf{Answer to RQ2:} The results demonstrate that CBD is significantly more computationally efficient than the GD metric. Across all input sets and subset sizes, CBD requires roughly \Rev{2.6 to 35} times less computation time while achieving comparable diversity measurement effectiveness. This finding confirms CBD as a practical and scalable diversity metric for large-scale DNN input selection.
\end{tcolorbox} 





\begin{table*}[b]
    \centering   
    \caption{Average accuracy improvement percentages achieved by fine-tuning the DNN model with input subsets selected by the CBD-based approach and each baseline (five runs)}
    \resizebox{0.9\textwidth}{!}{
    \begin{tabular}{|c|c|      c|     c|c|  c| c|  c| c|  c|}
    \hline 
    Input &Selection &CBD-based &\multicolumn{2}{c|}{\Revv{Ablation baselines}} &\multicolumn{5}{c|}{Baselines} \\ \cline{4-10}
    set   &size      &selecion   &Margin &\Revv{RS+CBD}  &DATIS   &RTS &DeepGD &SETS &\Rev{GTS} \\ \hline
    \multirow{6}{*}{CIFAR-10}  
    &$b$=3\%   &\textbf{67.48\%}      &58.12\%   &\Revv{27.93\%}    &32.72\%       &64.68\%         &66.17\%  &36.06\%  &\Rev{32.16\%} \\ \cline{2-10}
    &$b$=5\%   &\textbf{73.23\%}		&72.63\%  &\Revv{50.35\%}	 &51.94\%   &61.11\%       &54.98\%  &45.12\%  &\Rev{56.10\%}  \\ \cline{2-10}
    &$b$=7\%   &\textbf{72.61\%}	&53.76\%  &\Revv{40.08\%}	 &56.98\%  &66.4\%          &45.7\%  &49.85\%  &\Rev{61.41\%}  \\ \cline{2-10}
    &$b$=10\%  &\textbf{86.48\%}	&50.48\%  &\Revv{36.81\%}	 &35.45\%  &69.36\%         &55.24\%  &44.24\%  &\Rev{69.79\%}  \\ \cline{2-10}
    &$b$=15\%  &\textbf{82.06\%}	&50.42\%  &\Revv{32.49\%}	 &34.87\%   &55.8\%          &54.6\%  &50.49\%  &\Rev{75.17\%}  \\ \cline{2-10}
    &$b$=20\%  &\textbf{92.68\%}	&47.15\%  &\Revv{29.64\%}	 &28.69\%   &34.19\%         &67.98\%  &54.32\%  &\Rev{81.61\%}  \\ \hline
    
    \multirow{5}{*}{ImageNet}  
    &$b$=5\%   &\textbf{22.66\%}    &7.29\%    &\Revv{6.21\%}    &10.91\%     &19.68\%     &7.79\%     &9.37\%  &\Rev{12.61\%}  \\ \cline{2-10}
    &$b$=7\%   &\textbf{32.37\%}     &11.56\%   &\Revv{8.62\%}     &26.68\%    &23.04\%     &15.10\%      &24.26\%  &\Rev{25.36\%}  \\ \cline{2-10}
    &$b$=10\%  &\textbf{33.27\%}    &20.12\%  &\Revv{18.71\%}  &28.16\%     &11.63\%       &19.66\%    &30.72\%   &\Rev{29.78\%}  \\   \cline{2-10}
    &$b$=15\%  &\textbf{41.81\%}     &16.80\%  &\Revv{17.07\%}  &32.62\%     &18.84\%    &32.89\%        &39.52\%  &\Rev{33.01\%}  \\   \cline{2-10}
    &$b$=20\%  &\textbf{62.20\%}    &24.64\%   &\Revv{26.62\%}  &48.39\%     &38.84\%    &34.35\%       &37.53\%  &\Rev{46.72\%}  \\   \hline

    \multirow{5}{*}{\Rev{CIFAR-100}}  
    &\Rev{$b$=3\%}    &\Rev{\textbf{32.50\%}}    &\Rev{8.45\%}  &\Revv{12.55\%}   &\Rev{11.82\%}   &\Rev{4.39\%}    &\Rev{11.32\%}     &\Rev{11.92\%}  &\Rev{13.67\%}  \\ \cline{2-10}
    &\Rev{$b$=5\%}    &\Rev{\textbf{34.46\%}}   &\Rev{8.27\%}   &\Revv{9.54\%}    &\Rev{9.26\%}     &\Rev{6.22\%}   &\Rev{13.87\%}   &\Rev{29.25\%}  &\Rev{16.14\%}  \\ \cline{2-10}
    &\Rev{$b$=7\%}    &\Rev{\textbf{49.68\%}}   &\Rev{20.19\%}  &\Revv{19.44\%}     &\Rev{15.40\%}  &\Rev{10.53\%}   &\Rev{36.03\%}    &\Rev{30.97\%}  &\Rev{24.23\%}  \\ \cline{2-10}
    &\Rev{$b$=10\%}   &\Rev{\textbf{54.26\%}}   &\Rev{44.42\%}   &\Revv{24.80\%}    &\Rev{20.71\%}   &\Rev{20.34\%}     &\Rev{32.33\%}  &\Rev{39.47\%}  &\Rev{38.83\%}  \\ \cline{2-10}
    &\Rev{$b$=15\%}   &\Rev{\textbf{66.70\%}}    &\Rev{41.55\%}   &\Revv{32.31\%}     &\Rev{32.01\%}   &\Rev{33.98\%}    &\Rev{50.16\%}     &\Rev{61.91\%}  &\Rev{43.46\%}  \\ \cline{2-10}
    &\Rev{$b$=20\%}   &\Rev{\textbf{74.45\%}}   &\Rev{49.60\%}  &\Revv{44.48\%}      &\Rev{54.82\%}  &\Rev{48.39\%}   &\Rev{55.10\%}     &\Rev{58.54\%}  &\Rev{52.03\%}  \\ \hline

    \end{tabular}
    }
    \label{tab:AccuracyImprovements}
\end{table*}

\subsection{\textbf{RQ3: Effectiveness of CBD-based input selection in guiding DNN model accuracy improvement}}
\label{sec:ResultsRQ3}

To address this research question, we investigated the improvement in DNN model accuracy achieved by fine-tuning with input subsets selected by our CBD-based approach. We compared it with all selection baselines outlined in Section~\ref{sec:BackgroundInputSelection}. To ensure a fair comparison, we adopted the same configurations and parameter settings provided in the original implementation of each baseline. However, to apply DeepGD~\cite{aghababaeyan2024deepgd} on ImageNet, we reduced the number of generations in the NSGA-II algorithm to 10 because running NSGA-II with its default 300 generations on such a large input set is computationally prohibitive.

To perform our experiments, we first divided each subject's original test sets in Table~\ref{tab:Subjects} into two equal candidate and validation sets. For subjects with the ImageNet input set, we used a 70\% candidate and 30\% validation split, since ImageNet contains a large number of classes, requiring larger input subsets to effectively fine-tune deeper, more complex models such as ResNet-101. Subsequently, we perform our experiment by selecting inputs from the candidate set and measuring the accuracy of both the original and fine-tuned models on unseen inputs in the validation set.

We applied our CBD-based selection approach, along with all selection baselines outlined in Section~\ref{sec:BackgroundInputSelection}, to select input subsets with different sizes from the candidate set across all subjects. Then, we used each selected subset to fine-tune the subject DNN models by training for a few more epochs. 
Consequently, we can calculate the absolute performance improvement for each approach as the difference between the fine-tuned model's accuracy and the original model's accuracy on the unseen validation set. However, the absolute accuracy difference does not account for the maximum improvement achievable for a given subject model. To address this, we consider the maximum potential improvement achievable by fine-tuning the model on the entire candidate set. This value represents the upper bound on accuracy improvement, i.e., the accuracy gain achievable if all candidate inputs were labeled and used during fine-tuning.
Therefore, we calculate the accuracy improvement percentages for each subject and selection budget as follows:

\begin{equation}\label{Eq:Improvements}
\text{Imp} = \left( \frac{Acc_{fine-tuned} - Acc_{original}}{Acc_{max} - Acc_{original}} \right) \times 100
\end{equation}

\noindent where $Acc_{fine-tuned}$, $Acc_{original}$, and $Acc_{max}$ represent the accuracy of the fine-tuned model, original model, and the maximum accuracy of the model fine-tuned by the entire candidate set, respectively, all evaluated on the unseen validation set. Compared to calculating absolute improvements, this normalization formula enables a fair comparison across budgets and subjects.
To eliminate the impact of the inherent stochasticity of the fine-tuning process, we repeated fine-tuning for each selected subset five times with different random seeds and report the average accuracy improvement across runs.

Table~\ref{tab:AccuracyImprovements} presents the average accuracy improvement percentages achieved by the CBD-based approach and all baselines for all subjects and selection budgets. 
In this table, very small selection budgets in which subsets were selected by none of the investigated selection approaches were omitted, as these did not significantly improve the DNN model. For instance, for CIFAR-10 and ResNet-18, \Rev{and also CIFAR-100 and DenseNet121,} none of the subsets selected with a budget of $b$=1\% improved the model, whereas for ImageNet with the deeper, more complex \Rev{ResNet-101}, we remove subsets selected with budgets of $b$=1\% and 3\%. This is because these subsets are too small to update a deep model's weights and thus cannot significantly improve its accuracy.

Table~\ref {tab:AccuracyImprovements} highlights the highest accuracy improvement for each subject and selection budget in bold. The results demonstrate that the CBD-based approach consistently outperforms all baselines across subjects and selection budgets. 
To evaluate the statistical significance of these improvement differences, we conducted Wilcoxon signed-rank tests comparing the improvements from the CBD-based approach with each baseline at a significance level of $\alpha$ = 0.05. The results show that all p-values are below 0.05, confirming that the CBD-based approach achieves statistically significant improvements. 
Moreover, the CBD-based approach achieves practically significant average accuracy improvements of 79\% on CIFAR-10, \Rev{52\% on CIFAR-100,} and 38\% on ImageNet across all selection budgets. 
These results correspond to an average gain of approximately \Rev{7.2} percentage points on CIFAR-10, \Rev{11.8 percentage points on CIFAR-100}, and \Rev{5.7} percentage points on ImageNet relative to the second-best-performing selection baselines across all selection budgets.
\TRR{C3.10}{To further evaluate the practical magnitude of these improvements, we computed the matched-pairs non-parametric effect size from the Wilcoxon signed-rank test. Across all budget sizes and runs, the effect size ranges from $0.86$ to $0.87$ on CIFAR-10, from $0.84$ to $0.87$ on CIFAR-100, and from $0.75$ to $0.87$ on ImageNet against all baselines. These values demonstrate large effect sizes, confirming that our approach consistently outperforms the baselines by a substantial margin across runs.}

\TRR{C3.5}{In addition to the SOTA baselines, we evaluate two ablation baselines to isolate the contribution of each selection component of the proposed approach, with the results reported in Table~\ref{tab:AccuracyImprovements}. The CBD-based approach, as described in Algorithm~\ref{Alg:CBD-basedSelection}, consists of two main selection components: uncertainty-based selection for the first $b_{init}$ inputs and diversity-based selection for the remainder. Therefore, we consider an uncertainty-only baseline (i.e., Margin) and a diversity-only baseline where the initial uncertainty-based selection is replaced with Random Selection (RS). The empirical results demonstrate that our CBD-based approach significantly outperforms both ablation baselines across all subjects and selection sizes, confirming that both selection components contribute to its superior effectiveness.}

It is important to note that across all the investigated selection approaches in our study, increasing the selection budget does not necessarily lead to a monotonic increase in the percentage improvement in accuracy. Since these results present the average accuracy improvement percentages across five fine-tuning runs, such fluctuations are unlikely to be due to stochastic effects. Instead, they primarily reflect catastrophic forgetting~\cite {toneva2018empirical}. As Toneva \textit{et al.}~\cite{toneva2018empirical} define it, a forgetting event occurs when an individual training example transitions from being correctly classified to incorrectly classified over the course of learning, even without a distributional shift. This dynamic results in non-monotonic performance trends as additional inputs or training iterations are introduced, since newly included or reweighted inputs can both reinforce and interfere with previously learned representations. 
In our experiments, we observed a similar behavior across all selection approaches, including our proposed CBD-based selection, suggesting that model updates, even on diverse subsets, repeatedly induce localized forgetting and relearning dynamics. Nonetheless, the CBD-based selection consistently outperforms the baselines across all selection budgets, demonstrating its strong generalizability.






\begin{tcolorbox}   
    \textbf{Answer to RQ3:} The CBD-based selection approach consistently outperforms all baselines\Revv{, including the two ablation baselines,} in guiding input selection for improving the DNN model across all subjects and selection budgets. Furthermore, CBD-based selection provides statistically significant accuracy improvements, achieving average accuracy increases of 79\%, \Rev{52\%,} and 38\% across all budget sizes for CIFAR-10, \Rev{CIFAR-100,} and ImageNet, respectively. 
\end{tcolorbox}


\begin{table*}[t]
    \centering   
    \caption{Selection time (in seconds) of the proposed CBD-based selection and each baseline}
    \resizebox{0.8\textwidth}{!}{
    \begin{tabular}{|c|c|   c|     c|c|  c| c|  c| c|  c|}
    \hline 
    Input &Selection &CBD-based  &\multicolumn{2}{c|}{\Revv{Ablation baselines}} &\multicolumn{5}{c|}{Baselines} \\ \cline{4-10}
    set   &size      &selecion    &Margin &\Revv{RS+CBD}  &DATIS   &RTS &DeepGD &SETS &\Rev{GTS} \\ \hline
    \multirow{6}{*}{CIFAR-10}  
    &$b$=3\%   &2      &2 &\Revv{1}  &5      &72  &959 &0.3  &\Rev{28} \\ \cline{2-10}
    &$b$=5\%   &2       &2  &\Revv{1}  &5      &72  &1,341 &0.5  &\Rev{28} \\ \cline{2-10}
    &$b$=7\%   &2     &2   &\Revv{1}  &5      &72  &1,966 &1   &\Rev{28} \\ \cline{2-10}
    &$b$=10\%  &2     &2   &\Revv{1}  &5      &72  &3,476 &2 &\Rev{28} \\ \cline{2-10}
    &$b$=15\%  &2     &2   &\Revv{2}  &5      &72  &5,293 &5   &\Rev{28} \\ \cline{2-10}
    &$b$=20\%  &3     &2   &\Revv{2}  &5      &72  &9,067 &12    &\Rev{28} \\ \hline
    
    \multirow{6}{*}{ImageNet}  
    &$b$=3\%   &71  &70 &\Revv{2}   &970  &73,080  &2,644 &67  &\Rev{1,797} \\ \cline{2-10}
    &$b$=5\%   &73 &70  &\Revv{4}  &970  &73,080  &4,960  &86  &\Rev{1,797} \\ \cline{2-10}
    &$b$=7\%   &76 &70  &\Revv{8}  &970  &73,080  &8,292  &254   &\Rev{1,797} \\ \cline{2-10}
    &$b$=10\%  &85 &70  &\Revv{16}  &970 &73,080  &25,984  &826  &\Rev{1,797} \\   \cline{2-10}
    &$b$=15\%  &111 &70  &\Revv{42}  &970 &73,080  &89,174  &2,796  &\Rev{1,797} \\   \cline{2-10}
    &$b$=20\%  &154  &70  &\Revv{85}  &970 &73,080  &233,744  &7,091  &\Rev{1,797} \\   \hline

    \multirow{6}{*}{\Rev{CIFAR-100}}  
    &\Rev{$b$=3\%}   &\Rev{10}    &\Rev{10} &\Revv{1}  &\Rev{124}     &\Rev{1,156}     &\Rev{1,247}    &\Rev{0.5}   &\Rev{331} \\ \cline{2-10}
    &\Rev{$b$=5\%}   &\Rev{10}     &\Rev{10}  &\Revv{1} &\Rev{124}    &\Rev{1,156}     &\Rev{1,936}    &\Rev{1.2}  &\Rev{331} \\ \cline{2-10}
    &\Rev{$b$=7\%}   &\Rev{10}     &\Rev{10}  &\Revv{1} &\Rev{124}     &\Rev{1,156}  &\Rev{2,578}  &\Rev{3}   &\Rev{331} \\ \cline{2-10}
    &\Rev{$b$=10\%}  &\Rev{10}     &\Rev{10}  &\Revv{1} &\Rev{124}    &\Rev{1,156}  &\Rev{4,344}  &\Rev{4}    &\Rev{331} \\ \cline{2-10}
    &\Rev{$b$=15\%}  &\Rev{11}     &\Rev{10}  &\Revv{2} &\Rev{124}    &\Rev{1,156}  &\Rev{6,051}  &\Rev{7}   &\Rev{331} \\ \cline{2-10}
    &\Rev{$b$=20\%}  &\Rev{12}     &\Rev{10}  &\Revv{2} &\Rev{124}    &\Rev{1,156}  &\Rev{11,129}  &\Rev{15}    &\Rev{331} \\ \hline

    \end{tabular}
    }   
     \label{tab:time}
\end{table*}

\subsection{\textbf{RQ4: Efficiency of CBD-based input selection}}
\label{sec:ResultsRQ4}
To answer this research question, we empirically evaluate the computational efficiency of the proposed CBD-based selection approach by comparing its selection time with that of all baselines. Specifically, we measure the time required by each approach to select unlabeled input subsets of varying sizes from each subject’s candidate set. Table~\ref{tab:time} reports selection times (in seconds) for our CBD-based approach and each baseline when selecting subsets corresponding to 3\%, 5\%, 7\%, 10\%, 15\%, and 20\% of the candidate set.

As shown in Table~\ref{tab:time}, the selection time for input prioritization approaches, including uncertainty metrics such as Margin and DATIS, as well as RTS, remains constant across different selection budgets. This is because these approaches calculate a priority score for all unlabeled candidate inputs once and then rank them, producing an ordered list of inputs. Then, a subset can be selected from the top of the ordered list based on the required number of inputs, within the selection budget. As a result, their selection time is independent of the selection budget and depends solely on the time needed to rank all inputs in the candidate set.

In contrast, for selection approaches such as the proposed CBD-based selection and DeepGD, the selection time varies across different selection budgets. It is important to note that, while the proposed CBD-based approach first ranks all candidate inputs by their uncertainty score, as with prioritization approaches, it also iteratively calculates the CBD score for multiple candidate subsets until the specified selection budget is reached. Consequently, the total selection time of the CBD-based approach includes both the initial ranking time and the subsequent diversity calculation time.

It is important to note that to ensure a fair and consistent comparison, we report execution times in this table on a server equipped with an Intel\textregistered\ Xeon\textregistered\ Gold 6234 CPU (16 cores, 3.30 GHz), 192 GB of RAM, and an Nvidia Quadro RTX 6000 GPU with 24 GB of memory.

\begin{table}[b]
\centering
\caption{\Rev{Required time for initializing the CBD approach and baselines (seconds).}}
\resizebox{\columnwidth}{!}{
    \begin{tabular}{|c| c| c|  c|  c|  c|  c|  c|  }
    \hline
    \textbf{Subject} & \textbf{CBD-based} & \textbf{Margin} & \textbf{DATIS} & \textbf{RTS} & \textbf{DeepGD} & \textbf{SETS} & \textbf{GTS} \\
    \hline
    \Rev{CIFAR-10}   &\Rev{155}  &0 &\Rev{13}   &\Rev{293}   & 0 & 0 &\Rev{100}  \\ \hline
    \Rev{ImageNet}   &\Rev{3,916} &0 &\Rev{2,791} &\Rev{15,691} & 0 & 0 &\Rev{3,690} \\
    \hline
    \Rev{CIFAR-100}  &\Rev{236}  &0 &\Rev{65}   &\Rev{455}   & 0 & 0 &\Rev{180}  \\ \hline
    \end{tabular}
}
\label{tab:initializationTime}

\end{table}

The results in Table~\ref{tab:time} demonstrate that the CBD-based selection is highly efficient, requiring selection time close to that of simple output-probability-based uncertainty metrics, such as Margin, even on larger input sets. These selection times are significantly lower than those of the diversity-based and other hybrid baselines, such as DeepGD and RTS. For instance, selecting a subset with $b=20\%$ from ImageNet takes roughly 154 seconds using the CBD-based approach, whereas DeepGD and RTS require 233,744 and 73,080 Seconds, respectively. It is important to note that the reported selection time for DeepGD on the CIFAR-10 \Rev{and CIFAR-100} input sets is based on the default NSGA-II configuration of 300 generations. In contrast, for the ImageNet input set, we reduced this parameter to 10 generations, as running NSGA-II with the default 300 generations would require more than 80 days of computation, making it practically infeasible.

The results in Table~\ref{tab:time} further indicate that hybrid baselines are computationally expensive, primarily due to the high cost of diversity calculations. Although SETS~\cite{wang2025sets} is the least expensive among the hybrid baselines, it still requires substantially higher selection time than the proposed CBD-based selection and the neighbor-aware uncertainty-based baseline, DATIS~\cite{li2024distance}. In contrast, the CBD-based selection remains highly efficient, even on larger input sets like ImageNet, requiring significantly less time than DATIS while retaining the benefits of diversity-based selection.

While prioritization-based selection approaches, such as Margin and DATIS, as well as RTS, require the same time to select subsets of varying sizes (for the reasons discussed above), the selection time for all other approaches increases with the selection budget. However, this increase is substantially smaller for the CBD-based approach than for SETS and DeepGD. For instance, selecting a subset with $b$=10\% from ImageNet requires only about 14 more seconds than with $b$=3\% for the CBD-based approach, whereas DeepGD requires about 23,340 more seconds (about 10 times as much). 
This further confirms the scalability and practical efficiency of the CBD-based selection approach.

\TR{C2.4, C3.6}{In addition to the online selection time, the proposed approach and three of the baselines, namely DATIS, RTS, and GTS, require an initialization phase before input selection. Table~\ref{tab:initializationTime} reports the corresponding one-time initialization time for the CBD-based approach and these baselines. For the proposed CBD-based approach, initialization involves extracting embedding features from both the DNN and CLIP, training the linear aligner, and constructing the RCS. For other baselines, initialization mainly involves finding the noise threshold for RTS, extracting embedding features for DATIS and GTS, and training a GNN specifically for GTS.
As shown in Table~\ref{tab:initializationTime}, the initialization time ranges from 155 seconds for CIFAR-10 to 3,916 seconds for ImageNet. Although this cost is higher for our approach than that of DATIS, it is substantially lower than RTS across all evaluated subjects and is comparable to the recently proposed GTS approach. More importantly, this initialization is performed only once for each DNN model and input set and is therefore negligible when distributed over all subsequent input selection runs. In contrast, the online selection process is executed every time a new candidate input set is available and thus dominates the practical runtime of the approach. As shown in Table~\ref{tab:time}, the CBD-based selection approach requires execution times that are comparable to the simple Margin baseline, significantly lower than DATIS, RTS, DeepGD, and GTS, and remains practical even for large-scale input sets such as ImageNet. Consequently, while the proposed approach incurs a limited one-time initialization cost, it achieves highly efficient selection time, making it well suited for repeated deployment in practical input selection and DNN retraining scenarios.
}

\begin{tcolorbox}   
    \textbf{Answer to RQ4:} The CBD-based selection approach remains highly efficient with selection times close to simple uncertainty-based selections such as Margin. In contrast, diversity-based and hybrid baselines, such as RTS and DeepGD, require significantly more selection time, particularly on larger input sets such as ImageNet. 
    These results further highlight not only the practicality and computational advantage of the CBD-based approach but also its scalability, particularly for larger input sets.
\end{tcolorbox}

\Rev{\subsection{\textbf{RQ5: Optimal balance between uncertainty and diversity in the proposed hybrid input selection}} }
\label{sec:ResultsRQ5}

\TR{C1.7, C2.3}{
In this research question, we investigate how the size of the initial uncertainty-based selected subset ($b_{init}$) affects the effectiveness of the CBD-based approach. As described in Section~\ref{sec:CBD-basedSelectionApproach}, we first initialize the selected subset using the most uncertain inputs and then incrementally increase its diversity using the CBD metric. Consequently, $b_{init}$ balances uncertainty and diversity during selection. A smaller initial subset allows CBD to influence a larger portion of the selected inputs, whereas a larger initial subset places greater emphasis on uncertainty before CBD-based refinement begins.
}

\begin{figure}
    \centering
    
    \subfloat[CIFAR-100 with DenseNet121.]{
        \includegraphics[width=\columnwidth]{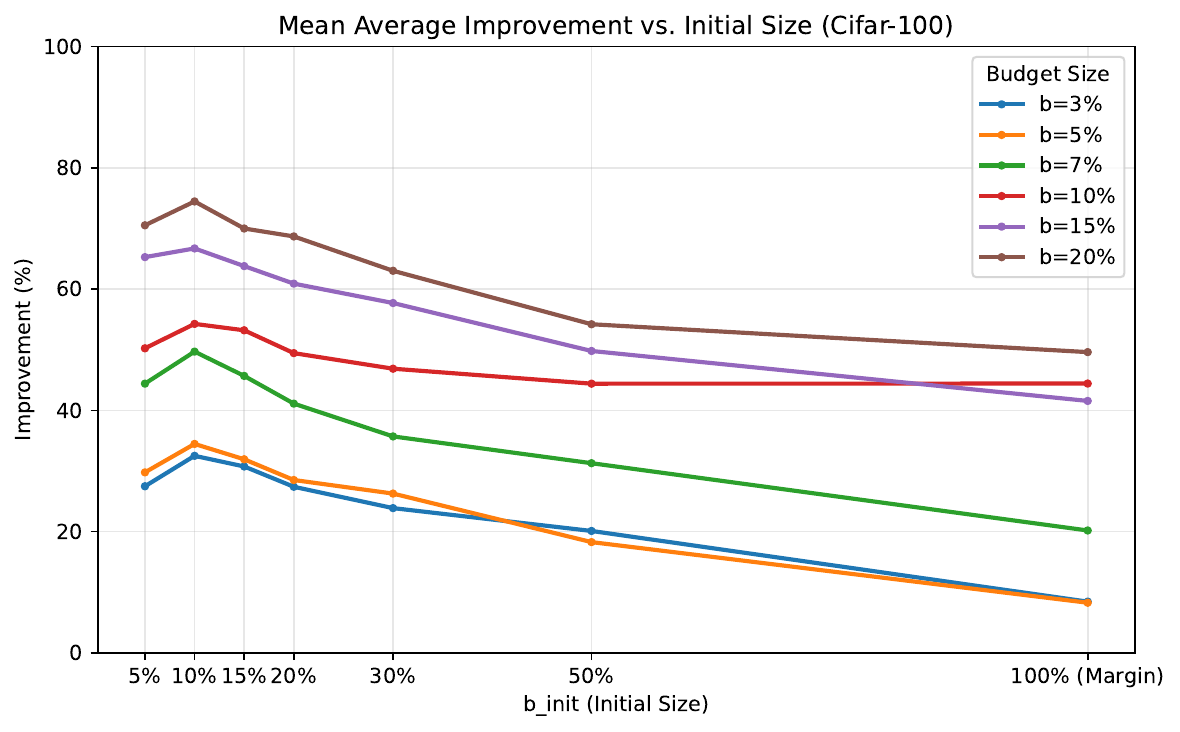}
        \label{fig:parameterAnalysis_cifar100}
    }
    
    \vspace{0.2cm}
    
    \subfloat[ImageNet with ResNet101.]{
        \includegraphics[width=\columnwidth]{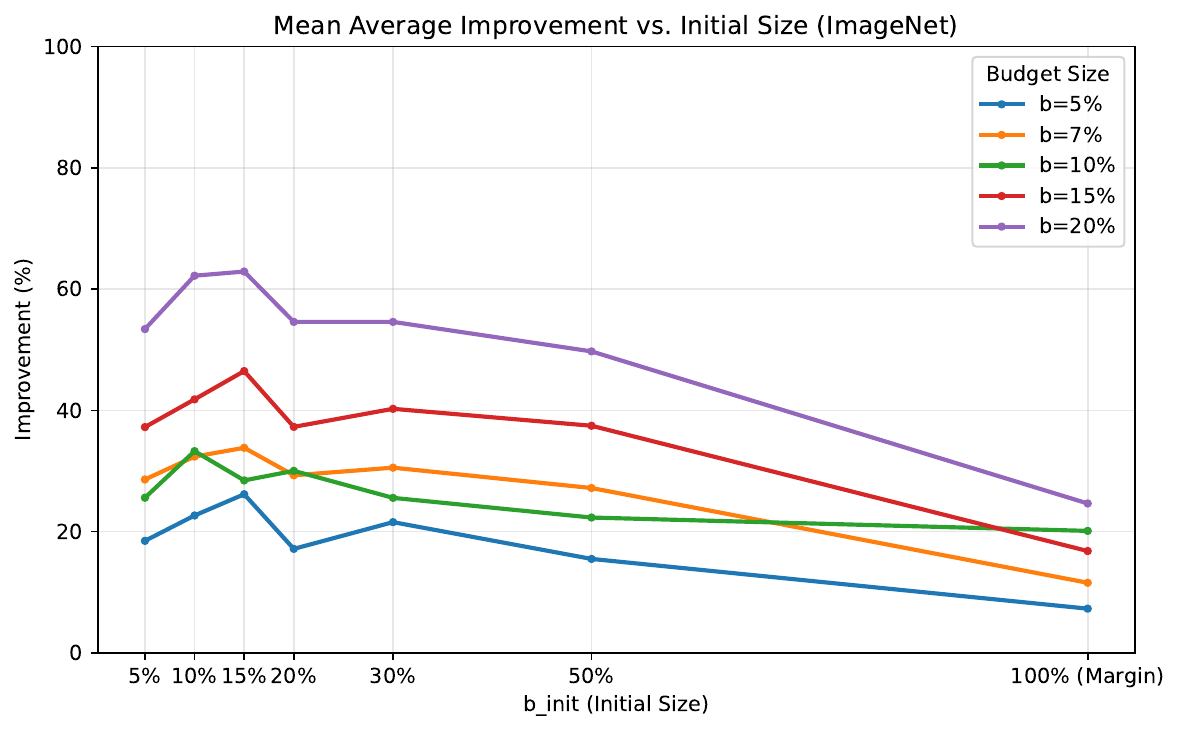}
        \label{fig:parameterAnalysis_imagenet}
    }

    \caption{The average accuracy improvement percentages achieved by the CBD-based approach with different parameter configurations ($b_{init}$).}
    \label{fig:parameterAnalysis}
\end{figure}

\TRR{C3.11}{To evaluate the effect of this parameter, we repeated our experiments 
varying $b_{init}$ from $0.05b$ to $b$, $b$ being the selection budget. Specifically, we considered alternative initial selection sizes ($b_{init}$) of $0.05b$, $0.1b$, $0.15b$, $0.2b$, $0.3b$, $0.5b$, and $b$}
for all investigated selection budgets ($b$) of 1\%, 3\%, 5\%, 7\%, and 10\% of the original test set and retrained the DNN model with the selected subset to report the accuracy improvements. Given the inherent stochasticity of the retraining process, we retrained the original DNN models five times using the selected subset for every single parameter variation and selection budget. Given the large number of parameter configurations evaluated, this experiment is exceptionally computationally intensive, requiring 30 input selection processes and 150 subsequent DNN retraining processes for each subject. 
\Revv{Therefore, to keep the analysis computationally feasible while ensuring representativeness, we perform the experiment on two of our subjects: ImageNet with ResNet101 and CIFAR-100 with DenseNet121.}
We report the results in Figure~\ref{fig:parameterAnalysis}.


\Rev{The results show that across all evaluated values, initializing the selected subset with approximately $5\%$ to $15\%$ of the selection budget consistently yields an optimal balance between uncertainty and concept-based diversity, achieving the highest average improvement. 
These results highlight the complementary roles of Margin and CBD in the proposed approach. When $b_{init}$ is very small, the initial subset may not contain enough highly informative inputs, reducing the effectiveness of the subsequent CBD-guided selection process. Conversely, increasing $b_{init}$ causes a larger portion of the selected subset to be determined solely by uncertainty. Since uncertainty metrics tend to prioritize inputs located near the decision boundary, many selected inputs exhibit similar characteristics, limiting the diversity of the final subset. Consequently, the effectiveness of the selected subset in improving the DNN gradually decreases as $b_{init}$ increases. 
While the greatest improvements are mostly achieved when $b_{init}$ is set to approximately $0.1b$, the proposed approach maintains its performance over a relatively broad range of configurations. These results indicate that selecting a small initial subset (between $0.05b$ and $0.15b$) based on uncertainty provides a better balance between uncertainty and diversity, allowing the CBD metric to effectively refine the remaining selected inputs. }

\begin{tcolorbox}   
    \textbf{\Rev{Answer to RQ5:}} 
    \Rev{Initializing the CBD-based selection with a small initial subset of approximately $0.05b$ to $0.15b$ provides a better balance between uncertainty and concept-based diversity, consistently yielding a high DNN model improvement. 
    }
\end{tcolorbox}




\subsection{Threats to Validity}
In this section, we discuss the different potential threats to the validity of our study and describe how we mitigated them.

\textbf{Construct threats to validity.}
A potential threat relates to the implementation of input selection baselines. To reduce the risk of implementation bias, we used the original implementations provided by the respective authors. Another threat concerns the accuracy improvement metric we used to evaluate the proposed approach. To mitigate this threat and provide a fair comparison between CBD-based selection and baselines across all subjects and selection budgets, we account for the maximum potential improvement achievable by fine-tuning the model on the entire candidate set in our accuracy improvement measurements. 


\textbf{Internal threats to validity.}
One internal threat to validity concerns the selection of parameters and experimental configurations for both the baseline and the proposed approach.
To mitigate this concern, we used the same settings provided in each baseline's original implementation. However, applying DeepGD \cite{aghababaeyan2024deepgd} to the ImageNet input set required a practical adjustment. Due to the prohibitive computational cost of running NSGA-II with its default 300 generations on such a large input set, we reduced the number of generations to 10.
The CBD-based selection approach relies on the parameter $m$, which controls how many concepts we extract from each image input to calculate the CBD score. As we described in Section~\ref{sec:ExperimentDesign}, we followed a simple manual procedure to determine an appropriate value for this parameter in our experiments. While the current results with $m=10$ are encouraging, optimizing this parameter may further improve CBD-based selection performance.
Another internal threat arises from the inherent randomness involved in fine-tuning DNN models using the selected inputs. We addressed this issue by repeating the fine-tuning process five times with different random seeds for each combination of selection budget, input set, and model, thereby reducing the impact of stochastic variations on our results.


\textbf{External threats to validity.}
To address potential concerns about the generalizability of the CBD-based selection approach, we conducted experiments across multiple subjects, covering diverse combinations of DNN models with varying internal architectures and input sets comprising different image types. In particular, we included large input sets such as ImageNet to evaluate the approach's scalability and practical applicability.
Additionally, the CBD-based selection approach was evaluated across a wide range of input selection budgets to assess its robustness under varying resource constraints. Across all combinations of input sets, model architectures, and selection budgets, the results consistently demonstrated substantial improvements in model accuracy, indicating strong generalizability across diverse experimental settings.

\Revv{Another potential threat concerns the robustness of the RCS and the CBD-based selection on unseen concepts.}
\TRR{C3.2}{While in any machine learning deployment setting, the target unlabeled inputs are assumed to share a reasonable degree of distributional and semantic alignment with the training inputs, natural semantic variation between the training inputs and unlabeled test inputs can introduce new unseen concepts. To investigate this threat, we conducted a comprehensive concept analysis across all subjects.
In our experiments, the concepts extracted from the training set (i.e., RCS) include 18,510 concepts for CIFAR-10, 33,366 for CIFAR-100, and 72,582 for ImageNet. We applied the same concept extraction procedure to the test sets, leveraging the complete Visual Genome knowledge base and directly using CLIP. Our analysis confirmed that the test sets indeed contain concepts absent from the corresponding RCS: 1,199 concepts (13\% of all concepts extracted from the test set) in CIFAR-10, 1,573 concepts (9\%) in CIFAR-100, and 138 concepts (1\%) in ImageNet. Despite the presence of these unseen concepts, the CBD-based approach achieves superior accuracy improvements, particularly on CIFAR-10 and CIFAR-100, which exhibit the highest proportions of unseen concepts. One possible explanation is that, for each image, we extract the top $m$ most similar concepts (with $m = 10$ in our experiments) and use the entire set to calculate the CBD score. Consequently, even if one or two concepts describing an image are missing from the RCS, the remaining highly related concepts can still provide a sufficiently accurate semantic representation for diversity estimation. These results provide empirical evidence that the CBD metric and CBD-based selection approach remain robust even when some of the unlabeled test inputs contain concepts not represented in the RCS.
}


\section{Related Work}
In this section, we first review existing work on input selection approaches for image DNN models. We then briefly summarize prior work that leverages VLMs to analyze and evaluate image classification DNNs.

\subsection{DNN input selection}
In recent years, several approaches to input selection have been proposed for DNN models. However, many of them are ineffective at improving the DNN model and are consistently outperformed by more recent approaches. For instance, several early contributions in input selection, such as NC~\cite{pei2017deepxplore, Ma2018DeepGaugeMT}, Cross-Entropy-based Sampling (CES)~\cite{li2019boosting}, and SA~\cite{kim2019guiding}, have been extensively studied in recent work. NC metrics assess inputs based on the extent to which neuron activation ranges are exercised. In contrast, SA metrics quantify how surprising an unlabeled input is with respect to the DNN’s training input distribution. CES prioritizes inputs based on the representations learned by the DNN under test: each input is characterized by the probability distribution of neuron outputs in the last hidden layer, and selection minimizes cross-entropy. All of these approaches, NC, AS, and CES, have been consistently outperformed by more recent input selection approaches, including several baselines in our experiments, such as RTS~\cite{sun2023robust} and DATIS~\cite{li2024distance}.

Uncertainty metrics are one of the most efficient selection approaches proposed for DNNs~\cite{hu2024test, Weiss2022SimpleTechniques, feng2020deepgini, li2024distance}. These metrics measure the DNN model's confidence in predicting an input and are used to prioritize selecting inputs for which the model is less confident, i.e., those it is more likely to mispredict. Simple uncertainty metrics such as DeepGini~\cite{feng2020deepgini}, Vanilla-Softmax~\cite{Weiss2022SimpleTechniques}, and Margin~\cite{hu2024test} are calculated solely based on the model's output probability vector~\cite{hu2024test}. Although DeepGini and Vanilla-Softmax have been widely used as baselines in prior work, including some of the baselines in our experiments, recent studies have consistently shown that they are outperformed by more advanced uncertainty-based and hybrid selection approaches~\cite{aghababaeyan2024deepgd, gao2022adaptive, sun2023robust, li2024distance}. However, SOTA uncertainty metrics such as DATIS~\cite{li2024distance} combine the model's output probability for a given input with information from its nearest neighbors to calculate the input's uncertainty score. 
Consequently, we also include Margin and DATIS as baselines. However, our proposed CBD-based selection approach outperformed both Margin and DATIS across all subjects and selection sizes. 

Diversity-based approaches are also among the well-studied and effective selection approaches for DNNs. Gao \textit{et al.} introduced ATS~\cite{gao2022adaptive}, which iteratively selects inputs that are less similar to the currently selected subset. This incremental selection strategy requires calculating the dissimilarity between each remaining input and the currently selected subset, making it one of the most computationally intensive selection approaches~\cite{hu2024test}. Moreover, our experiments consistently show that three baselines outperform ATS: DeepGD~\cite{aghababaeyan2024deepgd}, RTS~\cite{sun2023robust}, and DATIS~\cite{li2024distance}, both for fault detection and for improving DNN models. Aghababaeyan et al.~\cite {aghababaeyan2023black} have comprehensively investigated diversity metrics for image inputs, including Normalized Compression Distance, Standard Deviation, and GD. Their empirical results demonstrate that GD is the most effective metric for capturing the true diversity of an input set and, consequently, for guiding input selection to test and improve DNN models. However, to enhance the input selection performance, they combined the GD metric with DeepGini, a simple uncertainty metric, and proposed DeepGD~\cite{aghababaeyan2024deepgd}, a hybrid multi-objective selection approach.

In recent years, hybrid input selection strategies have received increasing attention, leading to the development of several effective approaches. These approaches primarily combine diversity and uncertainty metrics, leveraging their complementary strengths to select input subsets that are both diverse and highly effective at detecting faults, while also improving DNN performance through fine-tuning. Among these SOTA hybrid approaches, RTS~\cite{sun2023robust}, DeepGD~\cite{aghababaeyan2024deepgd}, SETS~\cite{wang2025sets}, and \Rev{GTS~\cite{Chen2026DiscrepantFeatures}} are included in our study as representative baselines.

A key limitation of these hybrid approaches lies in their computational cost. Calculating diversity for a large number of candidate subsets, combined with running multi-objective selection algorithms such as NSGA-II, can be highly resource-intensive. As our results demonstrate, while DeepGD and RTS are among the most effective approaches, they are computationally expensive and may even be impractical in extensive selection scenarios. SETS~\cite{wang2025sets}, proposed by Wang \textit{et al.}, attempts to mitigate this issue by incorporating a massive input reduction phase along with a highly restrictive selection process. Our results show that SETS substantially reduces the number of diversity calculations, making it less computationally expensive than RTS and DeepGD. \Rev{Similarly, GTS~\cite{Chen2026DiscrepantFeatures} is significantly more efficient than RTS and DeepGD, particularly on ImageNet. However, since it ranks all inputs at once, its selection time is generally higher than SETS for smaller budgets.}
However, the proposed CBD-based selection is highly efficient, requiring significantly less selection time than SETS.
Moreover, this reduction and restrictive selection of SETS also limit its effectiveness compared to the proposed CBD-based selection approach.

In contrast, our proposed CBD-based selection approach avoids discarding potentially valuable inputs and imposes no restrictive selection constraints. Instead, it significantly reduces the time required to compute diversity for each candidate subset by leveraging VLMs. As a result, the CBD-based approach achieves both high efficiency and high effectiveness, consistently outperforming all baselines, including existing hybrid methods, in improving DNN models.

In addition to uncertainty- and diversity-based input selection, some approaches move beyond purely heuristic metrics by exploiting structural or theoretical properties of DNN models. Multiple-Boundary Clustering and Prioritization (MCP)~\cite{shen2020multiple} clusters unlabeled inputs that lie near decision boundaries by leveraging the model’s output probability distributions. It then selects inputs evenly from these boundary-specific clusters, ensuring balanced coverage of decision regions while avoiding redundant selection from densely populated areas.
In contrast, CertPri adopts a different perspective by introducing a certifiable upper bound on the retraining gain of each unlabeled input. This bound is derived from a theoretical analysis of the expected reduction in training loss when incorporating a specific input during retraining. Nevertheless, both MCP and CertPri are outperformed by more recent approaches~\cite{li2024distance, sun2023robust, aghababaeyan2024deepgd}, including some of our baselines such as DATIS, RTS, and DeepGD.

Other input selection approaches, such as Mutation testing~\cite{humbatova2021deepcrime, hu2019deepmutation++} or other search-based selections~\cite{hao2023test} based on the NSGA-II algorithm, have also been proposed for DNN models. However, their huge computational cost limits their application in extensive selection scenarios. Consequently, we exclude these approaches from our study, since performance and scalability are our main objectives.


\subsection{DNN model analysis and testing using VLMs}
Since the introduction of VLMs, they have been leveraged to analyze and improve DNN models~\cite{abbasishahkoo2026cafd, hu2025debugging}. \Rev{For instance, a recent study introduced Concept-Aware Fault Detection (CAFD)~\cite{abbasishahkoo2026cafd}, which utilizes VLMs to estimate the likelihood that semantic concepts extracted from an input image are associated with DNN failures. By incorporating this complementary semantic information alongside the input's uncertainty score and its distance from its nearest training neighbors, CAFD achieves a significantly more effective fault detection rate.}
In another study, Hu \textit{et al.}~\cite{hu2025debugging} leveraged VLMs such as CLIP for runtime analysis of image classification DNN models and for localizing faults in their networks. They introduce semantic heatmaps to represent the frequency with which a specific concept is observed in correctly and incorrectly classified inputs from the model's training set. They then use these heatmaps to evaluate unlabeled inputs during runtime. Based on how similar a given input is to each heatmap, they can predict whether the model will misclassify or correctly classify the input and, finally, estimate the model's accuracy.
Furthermore, by performing zero-shot classification with a VLM, they identify whether misclassifications stem from faults in the DNN encoder or classification head. While the objectives of this approach differ from our goal, they share the basic idea of using VLMs to calculate the similarity between an image and a natural-language concept. However, a major limitation is that, to create concept heatmaps, they manually define a small, fixed set of concepts for each input class. Consequently, building heatmaps for input sets with many classes, such as ImageNet with 1000 classes, is not feasible.

\section{Conclusion}
In this paper, we introduce Concept-based Diversity (CBD), a new metric that effectively and efficiently measures the diversity of image input sets using Vision-Language Models (VLMs). Our evaluations show that the CBD score effectively captures the true diversity of images in an input set. More importantly, the CBD calculation is highly efficient, making it a scalable metric for repetitive, extensive input selection scenarios. Leveraging this metric, we propose a hybrid CBD-based input selection approach for DNN models that combines the model's uncertainty with the input set's diversity to select a small yet informative, representative subset of inputs for labeling and to improve the model through fine-tuning. We conducted an extensive empirical study to evaluate our CBD-based selection approach, comparing it against five highly effective SOTA baselines. The results show that the proposed approach consistently outperforms all baselines in accuracy while remaining among the most efficient, requiring computation times close to simple uncertainty metrics such as Margin. 

%

\section*{Acknowledgements}
This work was supported by the Canada Research Chair and Discovery Grant programs of the Natural Sciences and Engineering Research Council of Canada (NSERC) and the Research Ireland grant 13/RC/2094-2.


\bibliographystyle{IEEEtran}
\bibliography{main.bib}

\end{document}